\documentclass[11pt,letterpaper]{article}

\usepackage[margin=1in]{geometry}
\usepackage[T1]{fontenc}
\usepackage[utf8]{inputenc}
\usepackage{times}
\usepackage[authoryear,round]{natbib}
\setcitestyle{citesep={;},aysep={,},yysep={;}}
\usepackage{amsmath,amssymb,amsfonts,amsthm,mathtools}
\usepackage{algorithm}
\usepackage{algpseudocode}
\usepackage{booktabs}
\usepackage{array}
\usepackage{graphicx}
\usepackage{float}
\usepackage{placeins}
\usepackage{microtype}
\usepackage{xcolor}
\usepackage{colortbl}
\usepackage{hyperref}
\usepackage{url}
\hypersetup{
    colorlinks=true,
    linkcolor=black,
    citecolor=black,
    urlcolor=blue,
    pdftitle={What Should Federated LoRA Share? FedSAIL via Input-aware Subspace Alignment},
    pdfauthor={Junye Du, Shuaida He, Long Feng},
    pdfsubject={Federated low-rank adaptation},
    pdfkeywords={federated learning, LoRA, subspace alignment, personalized adaptation}
}

\newcommand{\R}{\mathbb{R}}
\newcommand{\E}{\mathbb{E}}
\newcommand{\tr}{\operatorname{tr}}
\newcommand{\diag}{\operatorname{diag}}
\newcommand{\TopEig}{\operatorname{TopEig}}

\newcommand{\I}{\mathbf{I}}
\newcommand{\one}{\mathbf{1}}
\newcommand{\bA}{\mathbf{A}}
\newcommand{\bB}{\mathbf{B}}
\newcommand{\bC}{\mathbf{C}}

\newcommand{\bG}{\mathbf{G}}
\newcommand{\bM}{\mathbf{M}}
\newcommand{\bP}{\mathbf{P}}
\newcommand{\bQ}{\mathbf{Q}}
\newcommand{\bR}{\mathbf{R}}
\newcommand{\bS}{\mathbf{S}}

\newcommand{\bV}{\mathbf{V}}
\newcommand{\bDelta}{\boldsymbol{\Delta}}
\newcommand{\bSigma}{\boldsymbol{\Sigma}}
\newcommand{\blambda}{\boldsymbol{\lambda}}
\newcommand{\bh}{\mathbf{h}}
\newcommand{\bv}{\mathbf{v}}
\newcommand{\calD}{\mathcal{D}}
\newcommand{\calL}{\mathcal{L}}

\newtheorem{remark}{Remark}
\title{\LARGE\bfseries What Should Federated LoRA Share? \\
FedSAIL via Input-aware Subspace Alignment}
\author{%
    Junye Du\thanks{Equal contribution.}\quad
    Shuaida He\footnotemark[1]\quad
    Long Feng\thanks{Corresponding author: \href{mailto:lfeng@hku.hk}{\texttt{lfeng@hku.hk}}.}\\[0.6em]
    {\normalsize School of Computing \& Data Science}\\
    {\normalsize The University of Hong Kong}\\[0.4em]
    {\small\href{mailto:junyedu@connect.hku.hk}{\texttt{junyedu@connect.hku.hk}}\quad
    \href{mailto:sdhe@hku.hk}{\texttt{sdhe@hku.hk}}\quad
    \href{mailto:lfeng@hku.hk}{\texttt{lfeng@hku.hk}}}%
}
\date{}

\begin{document}
\maketitle

\begin{abstract}
Federated low-rank adaptation (LoRA) requires identifying an update structure that is shared across heterogeneous clients. Prior work reports strong similarity among trained LoRA projection matrices across clients; however, such agreement may be largely induced by common initialization and collapses toward random overlap under independent initialization. 
More crucially, relying solely on parameter similarity inherently ignores the influence of local input regime.
To uncover a more robust shared structure, we introduce an input-aware action matrix that weights the adapter update by the second-moment statistics of local layer inputs. Empirically, while parameter similarity vanishes, the leading right singular directions of this action matrix remain strongly aligned across clients. This shared geometry preserves task-conditioned differences and naturally varies across network depths.
Motivated by these findings, we propose Federated Subspace-Guided Action-Informed Learning (FedSAIL). Instead of averaging weights, FedSAIL estimates a shared action subspace to regularize local training while preserving client-specific coefficients.
Across several benchmarks, our approach consistently improves predictive performance over competing federated LoRA methods while reducing communication cost significantly.
\end{abstract}

\section{Introduction}\label{sec:introduction}
\begin{figure}[t]
	\centering
	\includegraphics[width=\linewidth]{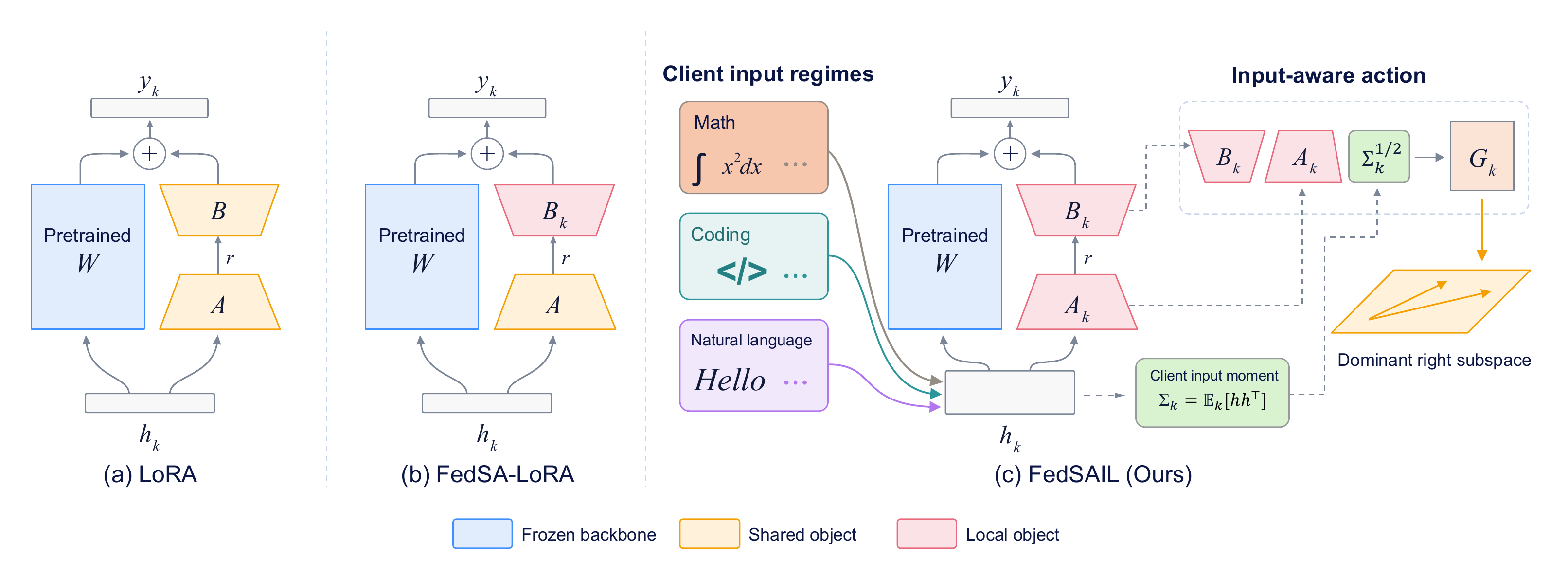}
	\caption{Sharing in federated LoRA.
		(a) LoRA.
		(b) FedSA-LoRA shares $\bA$ and retains $\bB_k$ locally.
		(c) FedSAIL shares a spectral summary of the dominant right subspace of $\bG_k=\bB_k\bA_k\bSigma_k^{1/2}$.}
	\label{fig:fedsail-overview}
\end{figure}

Low-rank adaptation (LoRA) fine-tunes a frozen pretrained layer through a compact update $\bDelta=\bB\bA$ \citep{hu2022lora}.  In federated learning, clients can exchange such adapters while keeping their data local \citep{mcmahan2017communication,zhang2024towards}.
A central question is therefore what information should be shared across clients. Existing methods address this by specifying which components of the adapter are communicated and how they are aggregated.
A representative example is FedSA-LoRA, which trains both factors but shares only $\bA$, while retaining client-specific $\bB_k$ for personalization \citep{guo2024selective}. 
For illustration, Figure~\ref{fig:fedsail-overview} (b) depicts the FedSA-LoRA mechanism, with the shared LoRA component highlighted in yellow.
Other approaches share $\bB$ against either a common frozen $\bA$ \citep{sun2024improving} or locally trained $\bA_k$ \citep{ban2026rethinking}. FLoRA aggregates full adapter updates by stacking factors \citep{wang2024flora}, while LoRA-FAIR refines aggregation and next-round initialization \citep{bian2025lora}.
FedRot-LoRA instead aligns factor coordinates through orthogonal rotations before aggregation, preserving each adapter product while addressing rotational misalignment \citep{zhang2026fedrot}.
These choices determine how clients represent and combine their updates, but parameter agreement alone need not identify structure learned in common.

Initialization provides one reason for this distinction. FedSA-LoRA motivates sharing $\bA$ by its higher cross-client similarity relative to $\bB$ \citep{guo2024selective}. However, recent analyses suggest that this apparent similarity may be largely inherited from common initialization \citep{zhu2024asymmetry,ban2026rethinking}. Our controlled RoBERTa/RTE \citep{liu2019roberta,wang2018glue} experiment further supports this interpretation: the rowspace agreement of the final $\bA$ falls to near-random levels under independent initialization, while the learned changes remain only weakly aligned under either initialization scheme.

A further limitation of sharing only LoRA parameters is that it does not explicitly account for client-specific input distributions. Since each client induces a local input regime at every adapted layer,  an update's effect depends jointly on its weights and the hidden states it encounters. Differences in domain, prompts, or sequence length can therefore make even identical adapters behave differently across clients.
These local input regimes are reflected in the hidden-state activations, motivating the use of activation statistics, such as second moments, to summarize their local geometry. 
Such statistics have already been used for adapter construction: EVA initializes LoRA input factors from dominant activation directions and allocates rank according to explained activation variance \citep{paischer2025eva}, while CorDA uses activation covariance to orient decompositions of pretrained weights \citep{yang2024corda}. 
While these methods use activation statistics to construct adapters, we instead ask whether such statistics can reveal shared structure in learned updates across client-specific input regimes, motivating a shift beyond raw parameter sharing.

To address this question, we introduce the input-aware action matrix $\bG_k=\bB_k\bA_k\bSigma_k^{1/2}$, where $\bSigma_k=\E_k\bigl[\bh\bh^\top\bigr]$ is the uncentered second moment of the local layer inputs. We defer its formal properties to Section~\ref{sec:action-matrix}, but intuitively, the leading right singular directions of $\bG_k$ identify the input subspace where the adapter's effect is most concentrated. This provides a natural and robust representation of cross-client common structure. Figure~\ref{fig:fedsail-overview} (c) contrasts this action-based perspective with existing parameter-sharing approaches.

Our empirical studies confirm that $\bG_k$ uncovers a robust shared structure often obscured by raw parameter comparisons. In our RoBERTa/RTE diagnostic, the leading action direction remains aligned across clients even under independent initialization and captures substantially more action energy on a held-out client than matched geometric controls.
Furthermore, a factorial GPT-2 study reveals that this geometric alignment is highly task-conditioned and depth-dependent: clients solving the same task remain more aligned across domains than clients solving different tasks. These results suggest a simple design principle: share only a compact common action subspace, retain client-specific flexibility, and use layerwise agreement to decide where collaboration is beneficial.

We realize this principle through \emph{Federated Subspace-Guided Action-Informed Learning} (FedSAIL). Instead of exchanging raw weights, each client uploads a compact spectral summary of its action matrix. The server then estimates a common geometric basis, which guides local training via a soft penalty on actions falling outside this shared subspace. Since both LoRA factors remain local, clients preserve their specific coefficients and trainable residual directions. Crucially, this design fundamentally decouples communication cost from local adapter capacity, allowing clients to scale up their models locally without incurring additional communication costs. Furthermore, by evaluating cross-client agreement after a local warmup phase, FedSAIL can restrict collaboration only to structurally aligned layers. Evaluated on personalized GLUE, FedSAIL consistently outperforms existing federated LoRA baselines. In highly heterogeneous domain-and-task settings, our layer selection strategy further reduces communication by $19.26\%$ while preserving comparable predictive accuracy, validating FedSAIL as a highly efficient and data-driven approach to federated fine-tuning.

We summarize our main contributions as follows:
\begin{itemize}
	\item We introduce an input-aware action matrix $\bG_k$ that exposes shared structure in federated LoRA beyond raw parameter similarity. Controlled diagnostics show that its leading directions remain aligned even under independent initialization, capturing structures that are inherently task-conditioned and depth-dependent.
	\item We develop FedSAIL, a novel federated fine-tuning framework that guides personalized adaptation via a shared action subspace. By keeping both LoRA factors strictly local and exchanging only compact spectral summaries, FedSAIL fundamentally decouples network communication overhead from local adapter capacity.
	\item We evaluate FedSAIL across heterogeneous client settings, including ablations over local rank and communicated summaries. Selective layer sharing further reduces communication while maintaining comparable mean accuracy in the studied settings.
\end{itemize}

\section{Identifying Shared Structure in Federated LoRA}\label{sec:what-to-share}
In this paper, we consider a frozen linear layer with input dimension $p$ and output dimension $q$. During local adaptation, client $k$ learns a rank-$r$ weight update parameterized as
\begin{equation}
    \bDelta_k=\bB_k\bA_k\in\R^{q\times p},\quad r<\min(p, q),
\end{equation}
where $\bA_k\in\R^{r\times p}$ and $\bB_k\in\R^{q\times r}$ denote the down-projection and up-projection matrices, respectively. For notational simplicity, we omit the layer index and the LoRA scaling factor.

Given a layer input $\bh\in\R^p$, the corresponding LoRA correction is $\bB_k\bigl(\bA_k\bh\bigr)$ \citep{hu2022lora}. This operation can be viewed as sequential feature-extraction and output-mapping: $\bA_k$ projects $\bh$ into an $r$-dimensional latent feature space, while $\bB_k$ maps the resulting features to a correction vector in $\R^q$ \citep{zhu2024asymmetry}. 
A fundamental question in federated LoRA then arises: what underlying  structure in this adaptation process is shared across clients,  and how can it be used to improve local
adaptation? 

\subsection{Limits of Raw Parameter Similarity}\label{sec:factor similarity}
To answer this question, we first examine the pitfalls of existing parameter-centric approaches. In federated learning, sharing a LoRA component typically means aggregating locally trained parameters across clients and broadcasting the result for the next round of local training. Existing methods share either the projection matrix $\bA_k$,  $\bB_k$, or the full adapter update $\bDelta_k$, each reflecting a different assumption about transferable structure. For example, sharing $\bA_k$ imposes a common input feature map while allowing client-specific $\bB_k$ to produce personalized output corrections.
Conversely, sharing $\bB_k$ imposes a common mapping from latent representation to the output space while retaining client-specific input projections \citep{sun2024improving}. As shown in Table~\ref{tab:sharing-mechanisms}, we summarize some shared structures and the components retained locally across existing methods.

\begin{table}[htbp!]
    \centering
    \caption{Comparison of shared components, aggregation strategies, and local training rules across different federated LoRA frameworks.}
    \label{tab:sharing-mechanisms}
    \footnotesize
    \setlength{\tabcolsep}{3.5pt}
    \renewcommand{\arraystretch}{1.0}
    \setlength{\arrayrulewidth}{0.3pt}
    \begin{tabular}{@{}
        >{\raggedright\arraybackslash}m{0.20\linewidth}
        >{\centering\arraybackslash}m{0.32\linewidth}
        >{\raggedright\arraybackslash}m{0.25\linewidth}
        >{\centering\arraybackslash}m{\dimexpr0.23\linewidth-6\tabcolsep\relax}
        @{}}
        \toprule
        \textbf{Method} & \textbf{Shared object}
        & \textbf{Aggregation} & \textbf{Local training} \\
        \midrule
        FedIT\newline{\scriptsize\citep{zhang2024towards}}
        & $\bA,\bB$
        & \shortstack[l]{$\bar{\bA}=\sum_k\omega_k\bA_k$\\[0.5pt]
            $\bar{\bB}=\sum_k\omega_k\bB_k$}
        & $\bA_k,\bB_k$ \\
        \arrayrulecolor{black!18}\specialrule{0.3pt}{1pt}{2pt}\arrayrulecolor{black}
        FLoRA\newline{\scriptsize\citep{wang2024flora}}
        & Adapter update $\bDelta$
        & $\bar{\bDelta}=\sum_k\omega_k\bDelta_k$
        & $\bA_k,\bB_k$ \\
        \arrayrulecolor{black!18}\specialrule{0.3pt}{1pt}{2pt}\arrayrulecolor{black}
        FFA-LoRA\newline{\scriptsize\citep{sun2024improving}}
        & $\bB$; common frozen $\bA^{(0)}$
        & $\bar{\bB}=\sum_k\omega_k\bB_k$
        & Only $\bB_k$ \\
        \arrayrulecolor{black!18}\specialrule{0.3pt}{1pt}{2pt}\arrayrulecolor{black}
        FedSA-LoRA\newline{\scriptsize\citep{guo2024selective}}
        & $\bA$
        & $\bar{\bA}=\sum_k\omega_k\bA_k$
        & $\bA_k,\bB_k$ \\
        \arrayrulecolor{black!18}\specialrule{0.3pt}{1pt}{2pt}\arrayrulecolor{black}
        FedDPA-LoRA\newline{\scriptsize\citep{yang2024dual}}
        & Global $\bA^g,\bB^g$
        & \shortstack[l]{$\bar{\bA}^g=\frac{1}{K}\sum_k\bA_k^g$\\[0.5pt]
            $\bar{\bB}^g=\frac{1}{K}\sum_k\bB_k^g$}
        & \shortstack{Global $\bA_k^g,\bB_k^g$\\[0.5pt]
            Local $\bA_k^\ell,\bB_k^\ell$} \\
        \arrayrulecolor{black!18}\specialrule{0.3pt}{1pt}{2pt}\arrayrulecolor{black}
        \textbf{FedSAIL} (ours)
        & \shortstack{Dominant right subspace\\[0.5pt]of $\bG_k$ in ~\eqref{eq:Gk}}
        & \shortstack[l]{$\widehat\bM=\sum_k\omega_k\widehat\bM_k$ in ~\eqref{eq:server-gram}\\[0.5pt]
            $\widehat\bV=\TopEig_s(\widehat\bM)$}
        & $\bA_k,\bB_k$ \\
        \bottomrule
    \end{tabular}
\end{table}
However, these parameter-sharing schemes do not by themselves reveal what structure is genuinely learned in common across clients.   As a representative example,  \citet{guo2024selective} motivate sharing $\bA_k$ based on its higher cross-client similarity relative to $\bB_k$. Yet, the factors in low-rank matrix factorizations are nonunique: $\bB_k\bA_k=\bigl(\bB_k\bR\bigr)\bigl(\bR^{-1}\bA_k\bigr)$ for any invertible $\bR\in\R^{r\times r}$. Consequently, directly sharing raw factor parameters constrains specific coordinate representations rather than coordinate-invariant subspaces.
Furthermore, prior studies suggest that high similarity among trained $\bA$ factors can arise from common initialization \citep{zhu2024asymmetry,ban2026rethinking}, which highlights the need to distinguish inherited similarity from genuine learned agreement.

\begin{figure}[htbp!]
    \centering
    \includegraphics[width=\linewidth]{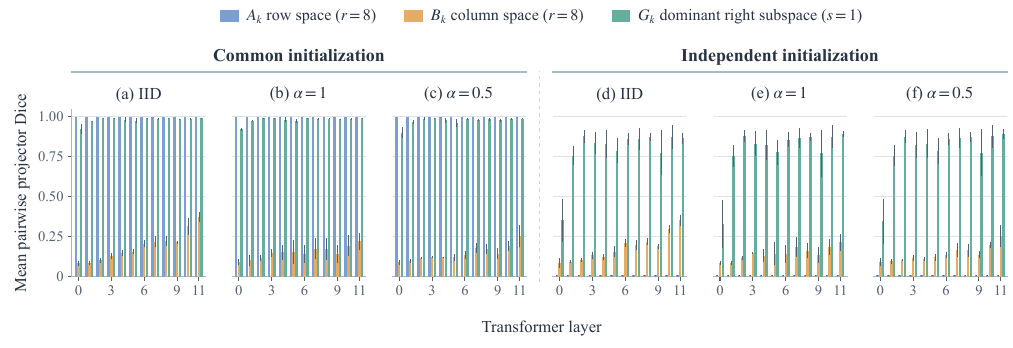}
    \caption{Cross-client subspace agreement for RoBERTa-base on RTE. For each initialization group, label heterogeneity increases from left to right, with $\alpha\in\{0.5,1.0\}$ controlling Dirichlet label skew.}
    \label{fig:shared-object}
\end{figure}

As an illustration, we compare common and independent initializations in a controlled RoBERTa-base study on RTE. We examine both the trained factors and their learned changes to assess whether clients develop a common representation during early adaptation. 
To rigorously quantify this cross-client geometric alignment, we employ the basis-invariant \emph{Projector Dice}. For two orthogonal projectors $\bP_1$ and $\bP_2$ of ranks $r_1$ and $r_2$, it is defined as:
\begin{equation}
\operatorname{Dice}(\bP_1,\bP_2) := \frac{2\tr(\bP_1\bP_2)}{r_1+r_2},
\end{equation}
which yields a value of one for perfectly identical subspaces and zero for completely orthogonal ones. Evaluated under this metric, our comparison reveals that high trained-$\bA$ similarity need not indicate genuinely shared learning. Figure~\ref{fig:shared-object} illustrates this pattern across layers and heterogeneity levels.
In both conditions, the final row spaces remain close to their respective initializations, while the learned changes are weakly aligned across clients. Thus, within this diagnostic, the apparent agreement primarily reflects inherited initialization geometry and is insufficient evidence of a common representation learned through adaptation. We refer to Appendix~\ref{app:real-diagnostics} for the detailed experimental setting and numerical results.

\subsection{Input-Aware Action Matrix}\label{sec:action-matrix}
If raw parameter similarity is an unreliable proxy for shared learning, where does the true cross-client shared component lie? We argue that the effect of a local update depends jointly on its weights and the hidden states it encounters. To characterize this, let $\bh\in\R^p$ denote the hidden-state vector entering the adapted layer on client $k$, and let $\E_k$ denote the expectation under its local input distribution. Assuming $\E_k|\bh|_2^2<\infty$, define $ \bSigma_k=\E_k\bigl[\bh\bh^\top\bigr]$. Empirical version of $\bSigma_k$ has been used to guide model merging \citep{jin2022dataless, salami2025closed}, weight pruning \citep{frantar2023sparsegpt}, and model compression \citep{wang2025svd}, revealing its potential for shaping parameter effects. Here, we define the input-aware action matrix
\begin{equation}\label{eq:Gk}
	\bG_k:=\bDelta_k\bSigma_k^{1/2}.
\end{equation}
Note that when $\bSigma_k=\I_p$, \eqref{eq:Gk} reduces to $\bG_k=\bDelta_k$. More generally, if $\bSigma_k=c_k\I_p$ for some $c_k>0$, the right singular subspaces of $\bDelta_k$ are preserved. In contrast, anisotropic input moments can alter both the orientation and relative strength of these singular directions, revealing shared structure that may be hidden in the raw updates.Crucially, this representation evaluates update differences strictly through their effects on the local input distribution. For any alternative update $\widetilde\bDelta$, we have:
	\[
	\bigl\|(\bDelta_k-\widetilde\bDelta)\bSigma_k^{1/2}\bigr\|_F^2
	=
	\E_k\bigl[
	\bigl\|(\bDelta_k-\widetilde\bDelta)\bh\bigr\|_2^2
	\bigr].
	\]
Thus,  update differences are weighted by the local input distribution, where directions carrying little input energy are naturally discounted.
Moreover, since $\bG_k$ depends on the LoRA factors only through their product $\bB_k\bA_k$, it is invariant to any invertible reparameterization of the raw factors.

Rather than synchronizing the entire action matrix, we can isolate its most critical geometric components. Let $\bV_{k,s}$ contain the leading $s$ right singular vectors of $\bG_k$, and define $\bP_{k,s}=\bV_{k,s}\bV_{k,s}^{\top}$.
Among rank-$s$ orthogonal projectors, $\bP_{k,s}$ maximizes $\|\bG_k\bP\|_F^2$, thereby identifying the low-dimensional subspace where the adapter's effect is most concentrated. This projector naturally provides a basis-invariant target for cross-client alignment. By employing guidance toward this subspace,  clients are able to preserve their personalized coefficients and residual directions without the constraints of rigid parameter averaging. To demonstrate how this action-based approach differs from traditional parameter-based views, the following remark shows that the nearly orthogonal LoRA rowspaces can yield an identical right action subspace after accounting for client-specific input moments. Besides, in Appendix~\ref{app:action-example}, we provide an intuitive geometric illustration.

\begin{remark}[Shared action geometry allows misaligned LoRA rowspaces]\label{rem:demo}
    Mathematically, nearly orthogonal LoRA rowspaces can induce identical right action subspaces after accounting for client-specific input moments. As a simple rank-one example, let $\bB_1=\bB_2=1$, $\bA_1=\bigl(a^{-1},1\bigr)$ and $\bA_2=\bigl(1,a^{-1}\bigr)$, with $\diag\bigl(a,1\bigr)$ and $\diag\bigl(1,a\bigr)$ for $a>1$. Then $\bG_1=\bG_2=\bigl(1,1\bigr)$, while the Dice similarity between the raw LoRA rowspaces converges to zero as $a\to\infty$. Appendix~\ref{app:action-example} provides an intuitive illustration.
\end{remark}

Beyond this illustrative construction, our empirical diagnostic with rank-$8$ LoRA adapters also reveals that cross-client agreement is concentrated in the dominant rank-$1$ right action subspace, while the remaining directions exhibit substantially weaker alignment. This observation
motivates sharing a compact geometric subspace while retaining greater local capacity for personalization. Detailed experimental evidence is provided in Appendix~\ref{app:dominant-components}.

\subsection{Task and Layer Dependence of Shared Structure}\label{sec:task-structure}
Beyond robustness to initialization, a meaningful shared component should also reflect the actual learning objectives of the clients, e.g., clients solving the same task should exhibit stronger geometric alignment than those solving different tasks. Furthermore, this alignment should naturally vary across the network's depth, as deeper layers typically specialize in task-specific representations. 

\begin{figure}[htbp!]
	\centering
	\includegraphics[width=\linewidth]{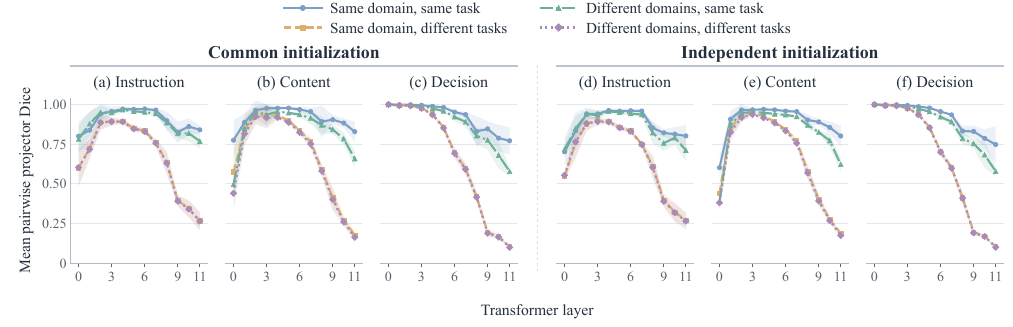}
	\caption{Task and layer dependence of action-subspace agreement in
    GPT-2. Curves report rank-$1$ Projector Dice averaged over query and
    value projections, while shaded bands indicate $95\%$ confidence intervals.}
	\label{fig:gpt2-task-structure}
\end{figure}
To verify this, we evaluate the action-subspace agreement in a controlled GPT-2 study, crossing two input domains (movie vs.\ product reviews) with two distinct classification objectives (sentiment vs.\ tense).  As shown in Figure~\ref{fig:gpt2-task-structure}, under both initialization schemes, same-task clients across domains exhibit stronger dominant-direction agreement on average than different-task clients
within the same domain. Besides, this distinction varies with depth: early Decision layers remain highly aligned across tasks, whereas deeper Content and Decision layers show substantially lower agreement between different-task clients. These observations complement our initialization analysis in Section~\ref{sec:factor similarity}. Detailed diagnostics are provided in Appendix~\ref{app:task-layer-diagnostics}.

\section{FedSAIL}\label{sec:method}

In Section~\ref{sec:what-to-share}, we demonstrate that parameter similarity alone is an unreliable proxy for client agreement, whereas the action geometry retains a robust and task-conditioned structure. Motivated by these, we propose \emph{Federated Subspace-Guided Action-Informed Learning} (FedSAIL). Instead of directly averaging adapter weights, FedSAIL extracts and shares the dominant right subspace of the input-aware action matrix. This allows the server to compute a basis-invariant geometric consensus to regularize local training, while keeping both LoRA factors and client-specific residual directions strictly local.

\subsection{Local Action Subspace Extraction}
To capture input-conditioned action geometry while keeping raw data
local, in our approach, each client $k$ estimates the uncentered second moment of its layer-input hidden states. After that, we apply shrinkage toward a scaled
identity matrix to improve the conditioning of this estimation and
stabilize subsequent spectral computations:
\begin{equation}
    \bS_k=\frac{1}{n_k}\sum_{i=1}^{n_k}\bh_{k,i}\bh_{k,i}^\top,
    \qquad
    \widehat\bSigma_k=(1-\gamma)\bS_k+
    \gamma\frac{\tr(\bS_k)}{p}\I_p.
    \label{eq:sigma-estimator}
\end{equation}
where $n_k$ is the number of local layer-input hidden states and $\gamma \in (0,1)$ determines the shrinkage level. Since constructing and sharing the full $q \times p$ action matrix $\widehat\bG_k$ is computationally inefficient, we instead form a compact equivalent:
\begin{equation}
    \widehat\bC_k=
    (\bB_k^\top\bB_k)^{1/2}\bA_k\widehat\bSigma_k^{1/2}
    \in\R^{r_{\rm L}\times p}.
    \label{eq:compact-c}
\end{equation}
Since $\widehat\bC_k^\top\widehat\bC_k
= \widehat\bG_k^\top\widehat\bG_k$, the compact matrix
$\widehat\bC_k$ has the same right singular subspaces and singular
values as the input-aware action matrix $\widehat\bG_k$. Client $k$ then isolates the highest-energy geometric directions by computing an orthonormal basis $\bQ_k = \TopEig_s(\widehat\bC_k^\top\widehat\bC_k)$ for the leading rank-$s$ eigenspace, alongside its corresponding eigenvalues $\blambda_k \in \R_+^s$. By sharing this subspace representation, the server captures the dominant directions of the input-weighted actions without requiring clients to synchronize their raw weight matrices.

\subsection{Server Aggregation and Personalized Feedback}

Upon receiving the local uploads, aggregating these subspaces presents a geometric alignment problem: directly averaging eigenvectors is ill-posed due to sign and basis-rotation ambiguities. Our approach resolves this by reconstructing a stabilized, trace-normalized rank-$s$ action Gram matrix for each client, and then aggregating them strictly as positive-semidefinite (PSD) matrices:
\begin{equation}
    \widehat\bM_k = \frac{\bQ_k\diag(\blambda_k)\bQ_k^\top}{\one^\top\blambda_k+\varepsilon_0}, \qquad
    \widehat\bM = \sum_{k=1}^K\frac{n_k}{\sum_jn_j}\widehat\bM_k.
    \label{eq:server-gram}
\end{equation}
Here, $\varepsilon_0 > 0$ acts as a numerical stabilizer. The trace normalization is a critical design choice: it controls for varying top-$s$ action magnitudes across heterogeneous clients, ensuring that clients with larger absolute updates do not disproportionately dominate the geometric consensus. The server then extracts the top-$s$ eigenspace of the aggregated Gram matrix, $\widehat\bV = \TopEig_s(\widehat\bM)$, and broadcasts the global projector $\widehat\bP = \widehat\bV\widehat\bV^\top$.

Armed with this global geometric consensus, clients resume training their personalized adapters. Rather than overwriting local weights, FedSAIL introduces a penalty that regularizes the local update toward the shared subspace:
\begin{equation}
    \min_{\bA_k,\bB_k} \widehat\calL_k^{\rm task}(\bB_k\bA_k) + \frac{\beta}{2} \left\|\bB_k\bA_k\widehat\bSigma_k^{1/2}(\I_p-\widehat\bP)\right\|_F^2.
    \label{eq:local-objective}
\end{equation}
This objective function selectively penalizes action energy that falls outside the globally shared space. Coefficients operating within that space remain entirely client-specific. Furthermore, the finite penalty weight $\beta$ provides local adapters with the necessary flexibility to retain unshared, residual directions if they sufficiently reduce the local task loss. This aligns with our observation in Figure~\ref{fig:gpt2-task-structure}: while different tasks share a dominant direction, they still require task-conditioned geometric variations to optimize performance.

\begin{algorithm}[H]
\caption{FedSAIL: Federated Subspace-Guided Action-Informed Learning}
\label{alg:fedsail}
\small
\begin{algorithmic}[1]
\Require Clients $\{\calD_k\}_{k=1}^K$; LoRA rank $r_{\rm L}$; shared rank
$s$; shrinkage $\gamma$; penalty $\beta$; stabilizer $\varepsilon_0$; rounds
$T$.
\ForAll{clients $k$ in parallel}
    \State Estimate $\widehat\bSigma_k$ by \eqref{eq:sigma-estimator}; locally
    warm up $(\bA_k^0,\bB_k^0)$ with $\beta=0$.
\EndFor
\For{$t=1,\ldots,T$}
    \ForAll{clients $k$ in parallel}
        \State Form $\widehat\bC_k^{t-1}$ by \eqref{eq:compact-c}.
        \State $\bQ_k^t\gets
        \TopEig_s((\widehat\bC_k^{t-1})^\top\widehat\bC_k^{t-1})$;
        collect the corresponding eigenvalues in $\blambda_k^t$.
        \State Upload $(\bQ_k^t,\blambda_k^t,n_k)$.
    \EndFor
    \State Server forms $\widehat\bM^t$ by \eqref{eq:server-gram} and
    broadcasts $\widehat\bV^t\gets\TopEig_s(\widehat\bM^t)$.
    \ForAll{clients $k$ in parallel}
        \State Update $(\bA_k^t,\bB_k^t)$ using \eqref{eq:local-objective}
        with $\widehat\bP^t=\widehat\bV^t(\widehat\bV^t)^\top$.
    \EndFor
\EndFor
\Ensure Personalized adapters $\{\bB_k^T\bA_k^T\}_{k=1}^K$.
\end{algorithmic}
\end{algorithm}

\section{Experiments}\label{sec:experiments}
In this section, we evaluate FedSAIL to validate the effectiveness of sharing input-aware action subspaces for personalized adaptation under data heterogeneity. We begin by comparing its predictive performance with existing federated LoRA methods. We then study how local adapter rank and shared subspace rank could affect accuracy and communication cost, before examining the contribution of $\bG_k$ sharing through ablation studies. Finally, we evaluate whether layer selection based on cross-client subspace agreement can reduce communication while retaining comparable predictive performance.

\subsection{Personalized Adaptation on GLUE}
\label{sec:glue-main}
We first evaluate RoBERTa-large on five GLUE benchmarks (RTE, SST-2, QNLI, MNLI, and QQP). To simulate a heterogeneous federated environment, we distribute each task across three clients using a Dirichlet label skew ($\alpha \in \{0.5, 0.75, 1.0\}$) and initialize each client's adapters independently. We compare FedSAIL against standard LoRA \citep{hu2022lora}, FFA-LoRA \citep{sun2024improving}, FedSA-LoRA \citep{guo2024selective}, and FedDPA-LoRA \citep{yang2024dual}. Full optimization details and hyperparameters are provided in Appendix~\ref{app:glue-protocol}.

Table~\ref{tab:glue-personalized} shows that FedSAIL achieves the highest
six-split mean validation accuracy at all three heterogeneity levels.
Its overall mean is $89.33\%$, compared with $88.69\%$ for FedSA-LoRA and
$88.57\%$ for FedDPA-LoRA, although the gains vary across tasks.
FFA-LoRA averages $74.62\%$ under the independent-initialization protocol.
This is a stress test of aggregating $\bB_k$ with different frozen
$\bA_k$ bases, rather than its canonical shared-$\bA$ setting
(Appendix~\ref{app:glue-protocol}). The results support input-aware
subspace guidance for personalized adaptation with independently
initialized adapters.

\begin{table}[htbp!]
    \centering
    \caption{Personalized GLUE validation accuracy (\%, mean $\pm$ sample std.) after $500$ rounds, with the highest mean in each column for each $\alpha$ shown in bold. Smaller $\alpha$ indicates stronger label heterogeneity.}
    \label{tab:glue-personalized}
    \small
    \setlength{\tabcolsep}{1.1pt}
    \begin{tabular*}{\linewidth}{@{\extracolsep{\fill}}clccccccc@{}}
        \toprule
        $\alpha$ & Method & RTE & SST-2 & QNLI & MNLI-m & MNLI-mm & QQP & Avg. \\
        \midrule
        $0.5$ & LoRA        & $78.70_{\pm 2.60}$ & $95.83_{\pm 0.93}$ & $90.13_{\pm 1.93}$ & $88.68_{\pm 1.60}$ & $88.96_{\pm 1.73}$ & $89.41_{\pm 5.75}$ & $88.62_{\pm 1.14}$ \\
              & FFA-LoRA    & $70.04_{\pm 5.91}$ & $90.75_{\pm 1.78}$ & $79.46_{\pm 2.06}$ & $65.46_{\pm 9.63}$ & $66.65_{\pm 9.11}$ & $81.28_{\pm 11.53}$ & $75.61_{\pm 3.27}$ \\
              & FedDPA-LoRA & $78.22_{\pm 6.83}$ & $95.99_{\pm 1.20}$ & $91.62_{\pm 1.21}$ & $\mathbf{89.06}_{\pm 1.58}$ & $\mathbf{89.29}_{\pm 1.71}$ & $89.62_{\pm 5.76}$ & $88.97_{\pm 1.33}$ \\
              & FedSA-LoRA  & $79.90_{\pm 8.61}$ & $95.95_{\pm 1.13}$ & $91.76_{\pm 0.61}$ & $88.56_{\pm 1.72}$ & $88.92_{\pm 1.76}$ & $89.53_{\pm 5.63}$ & $89.10_{\pm 1.60}$ \\
              & FedSAIL      & $\mathbf{80.87}_{\pm 7.50}$ & $\mathbf{96.37}_{\pm 1.17}$ & $\mathbf{93.06}_{\pm 0.57}$ & $88.98_{\pm 1.54}$ & $89.07_{\pm 1.75}$ & $\mathbf{89.80}_{\pm 5.39}$ & $\mathbf{89.69}_{\pm 1.28}$ \\
        \addlinespace
        $0.75$ & LoRA        & $\mathbf{77.26}_{\pm 0.96}$ & $95.45_{\pm 0.35}$ & $90.76_{\pm 0.89}$ & $88.39_{\pm 1.43}$ & $88.62_{\pm 1.44}$ & $88.68_{\pm 4.49}$ & $88.19_{\pm 0.54}$ \\
               & FFA-LoRA    & $61.13_{\pm 3.47}$ & $89.22_{\pm 2.81}$ & $81.77_{\pm 2.42}$ & $62.33_{\pm 10.84}$ & $63.60_{\pm 10.45}$ & $81.53_{\pm 8.88}$ & $73.26_{\pm 1.81}$ \\
               & FedDPA-LoRA & $74.61_{\pm 0.91}$ & $95.49_{\pm 0.46}$ & $92.12_{\pm 0.17}$ & $\mathbf{88.81}_{\pm 1.44}$ & $88.83_{\pm 1.45}$ & $88.67_{\pm 4.71}$ & $88.09_{\pm 0.27}$ \\
               & FedSA-LoRA  & $76.65_{\pm 3.86}$ & $95.41_{\pm 0.41}$ & $92.34_{\pm 0.29}$ & $88.22_{\pm 1.58}$ & $88.42_{\pm 1.75}$ & $88.70_{\pm 4.87}$ & $88.29_{\pm 0.39}$ \\
               & FedSAIL      & $\mathbf{77.26}_{\pm 5.05}$ & $\mathbf{95.64}_{\pm 0.20}$ & $\mathbf{93.23}_{\pm 0.61}$ & $88.61_{\pm 1.53}$ & $\mathbf{88.84}_{\pm 1.60}$ & $\mathbf{89.05}_{\pm 4.69}$ & $\mathbf{88.77}_{\pm 0.56}$ \\
        \addlinespace
        $1.0$ & LoRA        & $79.78_{\pm 2.53}$ & $95.83_{\pm 0.69}$ & $92.20_{\pm 0.94}$ & $88.18_{\pm 1.25}$ & $88.41_{\pm 1.51}$ & $88.00_{\pm 2.33}$ & $88.73_{\pm 1.00}$ \\
              & FFA-LoRA    & $76.29_{\pm 4.22}$ & $87.88_{\pm 0.98}$ & $80.08_{\pm 1.36}$ & $61.48_{\pm 11.53}$ & $62.48_{\pm 10.99}$ & $81.69_{\pm 4.80}$ & $74.98_{\pm 4.05}$ \\
              & FedDPA-LoRA & $77.98_{\pm 2.89}$ & $95.80_{\pm 1.26}$ & $92.80_{\pm 0.55}$ & $\mathbf{88.40}_{\pm 1.32}$ & $\mathbf{88.80}_{\pm 1.70}$ & $88.18_{\pm 2.16}$ & $88.66_{\pm 1.11}$ \\
              & FedSA-LoRA  & $79.54_{\pm 4.35}$ & $95.83_{\pm 0.81}$ & $92.32_{\pm 0.66}$ & $88.10_{\pm 1.52}$ & $88.17_{\pm 1.60}$ & $88.18_{\pm 2.14}$ & $88.69_{\pm 1.31}$ \\
              & FedSAIL      & $\mathbf{81.95}_{\pm 3.61}$ & $\mathbf{96.25}_{\pm 1.10}$ & $\mathbf{93.28}_{\pm 0.41}$ & $\mathbf{88.40}_{\pm 1.40}$ & $88.52_{\pm 1.35}$ & $\mathbf{88.72}_{\pm 2.11}$ & $\mathbf{89.52}_{\pm 1.14}$ \\
        \bottomrule
\end{tabular*}
\end{table}

\subsection{Shared Rank vs. Local Rank}
\label{sec:asymmetric-capacity}

A key architectural advantage of FedSAIL is its ability to decouple the rank of the globally shared subspace ($s$) from the capacity of the local adapters ($r_{\rm L}$). Standard federated parameter-averaging methods usually tie these two dimensions together. In contrast, FedSAIL's asymmetric design allows us to disentangle communication cost from client-side learning capacity. Table~\ref{tab:glue-capacity-tradeoff} reports the performance and communication costs as we vary these ranks.

We first evaluate the communication-accuracy trade-off by fixing the local rank at $r_{\rm L}=8$ and varying the shared rank. Expanding the shared subspace from $s=1$ to $s=4$ quadruples the communication payload, yet yields only small, non-monotonic changes in accuracy. This confirms that sharing dominant action direction captures the most critical geometric alignment information. Then we fix the shared rank at the communication-efficient $s=1$ and scale up the local adapter rank $r_{\rm L}$ from 2 to 8, predictive performance steadily improves across all heterogeneity levels. Crucially, since the shared spectral message depends entirely on $s$, scaling up $r_{\rm L}$ enhances local personalization capacity without incurring any additional communication overhead.

\begin{table}[htbp!]
    \centering
    \caption{Sensitivity to shared and local ranks on personalized GLUE . We report averaged validation accuracy (\%, mean $\pm$ sample std.), adapter parameter counts, and total communication costs.}
    \label{tab:glue-capacity-tradeoff}
    \small
    \setlength{\tabcolsep}{6pt}
    \begin{tabular*}{\linewidth}{@{\extracolsep{\fill}}ccccccc@{}}
        \toprule
        $r_{\rm L}$ & $s$ & Params. (M) & Comm. (GiB)
        & $\alpha=0.5$ & $\alpha=0.75$ & $\alpha=1.0$ \\
        \midrule
        $8$ & $1$ & $0.786$ & $0.4949$ & $89.69_{\pm 1.28}$ & $88.77_{\pm 0.56}$ & $89.52_{\pm 1.14}$ \\
        $8$ & $2$ & $0.786$ & $0.9895$ & $89.83_{\pm 1.08}$ & $88.87_{\pm 0.49}$ & $\mathbf{89.65}_{\pm 0.98}$ \\
        $8$ & $4$ & $0.786$ & $1.9787$ & $\mathbf{89.91}_{\pm 1.48}$ & $89.15_{\pm 0.18}$ & $89.47_{\pm 1.03}$ \\
        \midrule
        $2$ & $1$ & $0.197$ & $0.4949$ & $88.74_{\pm 1.43}$ & $87.92_{\pm 0.55}$ & $88.65_{\pm 1.41}$ \\
        $4$ & $1$ & $0.393$ & $0.4949$ & $89.21_{\pm 1.51}$ & $88.24_{\pm 0.81}$ & $89.10_{\pm 1.09}$ \\
        $16$ & $1$ & $1.573$ & $0.4949$ & $89.84_{\pm 1.43}$ & $\mathbf{89.22}_{\pm 0.29}$ & $89.61_{\pm 0.95}$ \\
        \bottomrule
    \end{tabular*}
\end{table}

\subsection{Ablation Study: Input-Moment Weighting in the Shared Message}

To isolate the source of FedSAIL's empirical gains, we ablate the rank-$1$ spectral message uploaded by the clients. To evaluate these properties, we compare shared subspaces extracted from the raw input factor ($\bA_k$), the full raw update ($\bDelta_k=\bB_k\bA_k$), and the input-aware action matrix ($\widehat\bG_k=\bDelta_k\widehat\bSigma_k^{1/2}$). All variants use identical communication payloads and retain the same
$\widehat\bG_k$-based local regularizer. In Appendix~\ref{app:core-ablation}, we provide the protocol details and task-level results.

As shown in Table~\ref{tab:glue-message-ablation}, the input-aware $\widehat\bG_k$ message outperforms both
unweighted messages across all three heterogeneity settings. At fixed communication cost, input-moment weighting therefore improves the guidance provided by the
shared subspace. Although $\bDelta_k=\bB_k\bA_k$ is already invariant to
invertible LoRA reparameterizations, its unweighted message performs
poorly under the same action-based regularizer. This supports selecting
shared directions that reflect local input distributions even after
removing parameterization ambiguity. We shall note that in Appendix~\ref{app:core-ablation}, an additional ablation study reveals that extracting the subspace from $\bA_k\widehat\bSigma_k^{1/2}$ performs only marginally lower than the full $\widehat\bG_k$.  Therefore, we do not claim that including the output factor $\bB_k$ is strictly necessary for empirical performance improvements. However, in comparison to $\bA_k\widehat\bSigma_k^{1/2}$, the full action matrix $\widehat\bG_k$ is invariant to invertible LoRA reparameterizations, which is a desirable property for a shared geometric object.
\begin{table}[htbp!]
    \centering
    \caption{Ablation of the shared message on personalized GLUE. We report validation accuracy (\%, mean $\pm$ sample std.) with rank-$1$ payloads.}
    \label{tab:glue-message-ablation}
    \small
    \setlength{\tabcolsep}{6pt}
    \renewcommand{\arraystretch}{1.12}
    \begin{tabular*}{\linewidth}{@{\extracolsep{\fill}}lccccc@{}}
        \toprule
        Shared message & Input-aware & $\alpha=0.5$ & $\alpha=0.75$ & $\alpha=1.0$ & Overall \\
        \midrule
        $\bA_k$ & No & $68.75_{\pm 2.92}$ & $66.17_{\pm 1.41}$ & $68.95_{\pm 3.47}$ & $67.96_{\pm 1.20}$ \\
        $\bDelta_k=\bB_k\bA_k$ & No & $68.79_{\pm 3.05}$ & $66.17_{\pm 1.42}$ & $69.04_{\pm 3.41}$ & $68.00_{\pm 1.22}$ \\
        \addlinespace[2pt]
        $\widehat\bG_k=\bDelta_k\widehat\bSigma_k^{1/2}$ & Yes & $\mathbf{89.69}_{\pm 1.28}$ & $\mathbf{88.77}_{\pm 0.56}$ & $\mathbf{89.52}_{\pm 1.14}$ & $\mathbf{89.33}_{\pm 0.36}$ \\
        \bottomrule
    \end{tabular*}
\end{table}

\subsection{Layer selection and communication}
\label{sec:layer-selection}

As observed in Section~\ref{sec:task-structure}, under highly heterogeneous environments, such as when clients operate across different domains and different tasks (DDDT), cross-client geometric agreement tends to decline sharply in deeper layers. To exploit this property, we construct a DDDT benchmark with eight RoBERTa-large clients split across IMDb/Amazon (sentiment) and QQP/QNLI (paraphrase/entailment). After a local warmup phase, we evaluate the cross-client geometric agreement at each layer. If the subspace agreement falls below a threshold $\tau$, that layer is excluded from subspace sharing entirely and remains purely local.
\begin{table}[!htbp]
    \centering
    \caption{Layer selection performance on DDDT. We report averaged held-out accuracy (\%, mean $\pm$ sample std.) at round $64$ and total communication cost, including selector overhead.}
    \label{tab:dddt-summary}
    \small
    \begin{tabular*}{\linewidth}{@{\extracolsep{\fill}}lccc@{}}
        \toprule
        FedSAIL variant & Accuracy (\%) & Comm. (MiB) & Saving (\%) \\
        \midrule
        Full & $90.033\pm0.225$ & $96.094$ & -- \\
        Selection, $\tau=0.6$ & $90.133\pm0.038$ & $77.589$ & $19.26$ \\
        Selection, $\tau=0.7$ & $89.908\pm0.800$ & $58.904$ & $38.70$ \\
        \bottomrule
    \end{tabular*}
\end{table}

At $\tau=0.6$, the selector leaves the final six layers (zero-based layers 18--23) local across all three training seeds, without an imposed contiguity constraint. This pattern is consistent with the decline in cross-client agreement at deeper layers observed in Section~\ref{sec:task-structure}. Table~\ref{tab:dddt-summary} shows that restricting guidance to layers with greater action-subspace agreement reduces communication by $19.26\%$, including selector overhead, while retaining comparable mean held-out accuracy. A higher threshold further reduces communication, with a modest decrease in mean accuracy. These results support using action geometry to guide where clients share subspaces, allowing the depth of collaboration to adapt to observed agreement while preserving local training in unshared layers. Appendix~\ref{app:layer-selection} provides the selection rule, complete threshold sweep, and communication accounting.

\section{Conclusion}
In this paper, we demonstrate that high parameter similarity in federated LoRA often reflects common initialization rather than genuine task learning. By shifting the focus from weight space to input-aware action space, we identify a robust, task-conditioned structure that genuinely generalizes across heterogeneous clients. Leveraging this geometric insight, we introduce FedSAIL. Instead of enforcing rigid parameter synchronization, FedSAIL aligns clients through a basis-invariant spectral consensus, providing regularization while preserving local coefficients and residual directions. Extensive evaluations confirm that FedSAIL consistently outperforms existing federated LoRA baselines. Additionally, by measuring layer-wise subspace agreement, our framework can use warmup-based selection to suspend communication at deeper layers with low cross-client agreement, reducing payload costs significantly while retaining comparable mean predictive accuracy. Ultimately,  we establish that effective federated fine-tuning must be data-driven: by anchoring consensus in local input distributions, it ensures clients collaborate where tasks align and diverge where personalization is essential.

\subsection*{AI use statement}
We used generative AI tools to provide feedback on the experimental methodology
and to assist with interpreting results. We also used these tools to revise manuscript and figures. The authors take
responsibility for the final content, including AI-assisted text, analyses,
and figures.

\subsection*{Ethics statement}
This work focuses on algorithmic design in federated learning and utilizes standard, publicly available datasets. The research does not involve human subjects, sensitive personal information, or proprietary data. Consequently, this paper does not present any specific ethical issues or potential for negative societal impact.

\subsection*{Reproducibility statement}
Algorithm~\ref{alg:fedsail} details the core FedSAIL framework. To ensure full reproducibility, comprehensive experimental details—including dataset construction, client partitioning, hyperparameter configurations, and exact communication accounting—are provided in Appendices~\ref{app:real-diagnostics} through~\ref{app:layer-selection}. Furthermore, the core implementation of FedSAIL, along with numerical tests and baseline methods, is anonymously available at \url{https://anonymous.4open.science/r/FedSAIL-4A7C/}.

\bibliographystyle{plainnat}
\bibliography{reference}

\clearpage
\appendix

\section*{Overview of the appendices}
The appendices provide the construction, diagnostic protocols, and additional
results supporting the main text. 
They are organized as follows.
\begin{itemize}
    \item Appendix~\ref{app:action-example} provides further details for Remark~\ref{rem:demo}, showing how shared action geometry can coexist with misaligned LoRA rowspaces.
    \item Appendix~\ref{app:real-diagnostics} describes the RoBERTa and GPT-2
    diagnostics for initialization effects, concentration in dominant action
    directions, and task and layer dependence, including their controls and
    statistical summaries.
    \item Appendix~\ref{app:glue-protocol} specifies the client partitions,
    training settings, baseline implementations, and evaluation protocol for
    personalized GLUE.
    \item Appendix~\ref{app:core-ablation} reports the shared-message ablation,
    which varies the use of output-factor and input-moment weighting while
    keeping the local action penalty fixed.
    \item Appendix~\ref{app:rank-sensitivity} studies shared and local rank
    sensitivity, with task-level accuracy, adapter parameter counts, and
    communication accounting.
    \item Appendix~\ref{app:layer-selection} details the real-data DDDT
    layer-selection study, including the benchmark protocol, geometric
    selection rule, threshold sweep, and selector overhead.
    \item Appendix~\ref{app:code-resources} describes the anonymous code
    release, its reproducibility scope, and the computational resources.
\end{itemize}

\section{Details of Remark~\ref{rem:demo}}\label{app:action-example}
\begin{figure}[!ht]
	\centering
	\includegraphics[width=\linewidth]{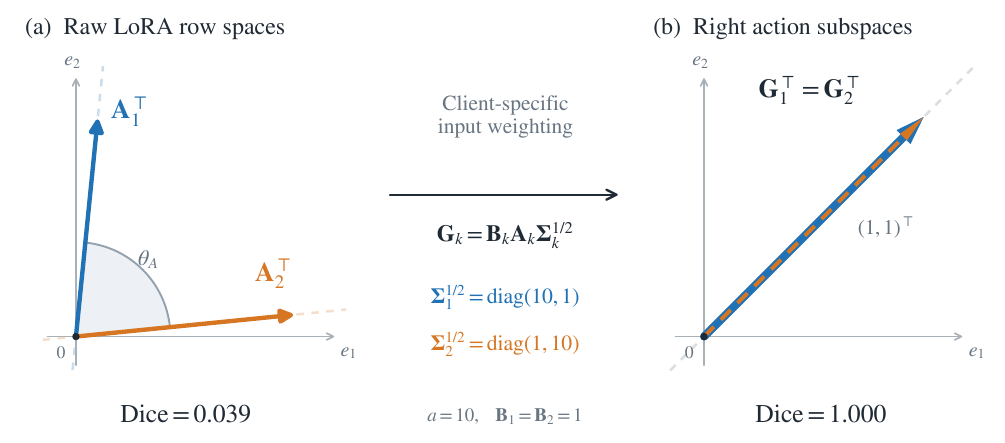}
	\caption{Illustration of Remark~\ref{rem:demo}, with $a=10$ and $\bB_1=\bB_2=1$. (a) The raw $\bA_k$ row spaces have Dice $0.039$. (b) After client-specific input weighting, $\bG_1=\bG_2=\bigl(1,1\bigr)$, so the right action subspaces coincide with Dice one. Blue and orange denote clients 1 and 2, respectively.}
	\label{fig:demo}
\end{figure}

The construction in Remark~\ref{rem:demo} relies on each adapter compensating for its client's input scaling.
With $\bSigma_1^{1/2}=\diag\bigl(a,1\bigr)$ and $\bSigma_2^{1/2}=\diag\bigl(1,a\bigr)$, we obtain
\[
    \begin{aligned}
    	\bG_1
    	&=\bigl(a^{-1},1\bigr)\diag\bigl(a,1\bigr)
    	=\bigl(1,1\bigr),\\
    	\bG_2
    	&=\bigl(1,a^{-1}\bigr)\diag\bigl(1,a\bigr)
    	=\bigl(1,1\bigr).
    \end{aligned}
\]
For each client, the smaller adapter coefficient exactly offsets the larger input scale in the corresponding coordinate. 
Both clients therefore have the same right action direction
$\bv=2^{-1/2}\bigl(1,1\bigr)^\top$. 
The corresponding projector $\bP=\bv\bv^\top$ satisfies
\[
	\bG_k\bigl(\I_2-\bP\bigr)=0,\qquad k=1,2,
\]
and hence captures all action energy for both clients. Despite this agreement, the raw LoRA rowspaces become increasingly misaligned.  Their Projector Dice is
\[
	\frac{\langle\bA_1,\bA_2\rangle_F^2}
	{\|\bA_1\|_F^2\|\bA_2\|_F^2}
	=
	\frac{\bigl(2a^{-1}\bigr)^2}
	{\bigl(1+a^{-2}\bigr)^2}
	=
	\biggl(\frac{2a}{a^2+1}\biggr)^2
	\to 0
\]
as $a\to\infty$.
Thus, the raw rowspaces are nearly orthogonal even though their input-weighted action directions remain identical. Figure~\ref{fig:demo} illustrates this contrast at $a=10$.

This example also clarifies the distinction between sharing an action subspace and sharing an input factor. 
For every $a>1$, the raw updates $\bDelta_1=\bA_1$ and $\bDelta_2=\bA_2$ are nonproportional. 
They cannot therefore be reproduced by a single shared rank-one factor $\bA$ with client-specific scalar coefficients $\bB_k$, since all products $\bB_k\bA$ would be proportional. A common action subspace thus accommodates both local updates under their respective input regimes, whereas exact input-factor sharing cannot represent them at the same rank.

\section{Additional Diagnostics of Shared Action Subspaces}
\label{app:real-diagnostics}

In this section, we supply the protocols and measurements behind
Section~\ref{sec:what-to-share}. We first examine initialization effects
in RoBERTa/RTE diagnostic, then compare
dominant and complete action subspaces, and finally separate task, domain,
and layer effects in GPT-2. The RoBERTa comparison isolates the choice of
adapter initialization under a fixed local-training protocol. These are
diagnostics of locally trained adapters; the federated
training evaluations are reported separately in Section~\ref{sec:glue-main}
and Appendices~\ref{app:glue-protocol}--\ref{app:layer-selection}.

\subsection{Experimental setup and measurement}
\label{app:diagnostic-protocol}

\paragraph{Training and initialization.}
Table~\ref{tab:diagnostic-protocol} summarizes the two studies. In both studies we freeze
the pretrained backbone and place LoRA on the query and value projections
in all 12 transformer layers. Common initialization uses identical initial
$\bA$ matrices at corresponding projections across clients; independent
initialization draws them separately. Both conditions initialize $\bB=0$.
Client partitions, optimization settings, minibatch orders, and training
dropout streams are held fixed across conditions. For RoBERTa, the dropout
seed is reset after injecting the initial adapters; the classification
head has a common initialization within each seed. Client 0 uses the same
initial $\bA$ in both conditions and serves as a paired anchor, not an
independent replicate. RoBERTa uses SGD without momentum, weight decay, or
gradient clipping. GPT-2 uses AdamW with zero weight decay and gradient
clipping at norm one.
\begin{table}[htbp!]
\centering
\caption{Local-training protocols for the structural diagnostics.}
\label{tab:diagnostic-protocol}
\small
\setlength{\tabcolsep}{5pt}
\begin{tabular}{lcc}
\toprule
Setting & RoBERTa-base / RTE & GPT-2 124M  \\
\midrule
Clients & 3 per partition setting & $2$ domains $\times$ $2$ tasks $\times$ $2$ replicas \\
Train / calibration / evaluation & $161$--$1543$ / $9$--$86$ / $10$--$87$ & $500 / 250 / 250$ \\
Local LoRA rank / scale & $8 / 16$ & $4 / 8$ \\
LoRA dropout & $0.05$ & $0$ \\
Steps / effective batch size & $200 / 128$ & $200 / 16$ \\
Maximum sequence length & $128$ & $64$ \\
Optimizer & SGD & AdamW \\
LoRA learning rate & $10^{-2}$ & $3\times10^{-4}$ \\
Classifier learning rate & $10^{-2}$ & No additional classifier \\
Training seeds & $123,124,125$ & $11,23,37,53,71$ \\
\bottomrule
\end{tabular}
\end{table}

We emphasize that the studies in this section is not to train high-performing models (full federated evaluations are deferred to Section~\ref{sec:experiments}). Instead, we simply want to examine how different adapter initializations affect local parameter updates across heterogeneous clients. 200 local steps without server aggregation are fully sufficient for this purpose.

\paragraph{RoBERTa client construction.}
We partition the 2,490 official RTE training examples across three clients using either IID shuffling or label-based Dirichlet allocation ($\alpha \in \{0.5, 1.0\}$). Under IID partitioning, each client receives exactly 830 examples. Under Dirichlet allocation, local dataset sizes vary from 161 to 1,543, with entailment ratios ranging between $8.67\%$ and $94.41\%$. The 277 official validation examples are distributed uniformly for IID clients and according to local training-label priors for Dirichlet clients. Each client's data is further stratified by label into calibration and evaluation subsets, retaining minor strata (e.g., the smallest calibration subset contains 9 examples from a single class).

\paragraph{GPT-2 tasks and token regimes.}
We evaluate controlled synthetic tasks using template-generated movie and product sentences across two classification objectives: sentiment and tense. For a given domain and replica, both tasks share identical underlying content but differ in instructions and labels. Each prompt comprises a task instruction ending in \texttt{Text:}, the content sentence, and an \texttt{Answer:} suffix. Training minimizes cross-entropy over the two candidate answer-token logits at the final prompt position, without appending answer tokens to the input. We divide each prompt into three disjoint diagnostic masks: \emph{content} (sentence tokens), \emph{decision} (the final prompt token predicting the answer), and \emph{instruction} (all remaining tokens, including non-decision suffix tokens). Held-out examples are jointly stratified by sentiment and tense into calibration and evaluation splits.

\subsection{Initialization and factor agreement}
\label{app:initialization-evidence}
\begin{table}[!ht]
\centering
\caption{Cross-client projector Dice in RoBERTa-base/RTE averaged across client pairs, layers, and projections. All entries use the fixed step-200 checkpoint.}

\label{tab:roberta-dominant}
\small
\setlength{\tabcolsep}{5pt}
\begin{tabular}{llccccc}
\toprule
Partition & Initialization & $\bA$ & $\bB$ & $\bG$: top 1 & $\bG$: top 2 & $\bG$: full \\
\midrule
IID & Common & $1.000$ & $0.199$ & $0.978$ & $0.825$ & $0.876$ \\
$\alpha=1$ & Common & $1.000$ & $0.185$ & $0.974$ & $0.834$ & $0.871$ \\
$\alpha=0.5$ & Common & $1.000$ & $0.177$ & $0.965$ & $0.818$ & $0.835$ \\
\midrule
IID & Independent & $0.011$ & $0.193$ & $0.814$ & $0.452$ & $0.159$ \\
$\alpha=1$ & Independent & $0.011$ & $0.180$ & $0.810$ & $0.454$ & $0.160$ \\
$\alpha=0.5$ & Independent & $0.011$ & $0.177$ & $0.810$ & $0.455$ & $0.159$ \\
\bottomrule
\end{tabular}
\end{table}

Following \citet{guo2024selective}, here we evaluate structural alignment using both the flattened-factor cosine similarity and basis-invariant subspace agreement (Table~\ref{tab:roberta-dominant}). Under common initialization, the flattened cosine for the final matrix $\bA$ reaches $1$, whereas under independent initialization, it drops to $-0.0003$ (with the corresponding $\bB$ cosines at $0.2238$ and $-0.0016$, respectively). Similarly, the final-$\bA$ rowspace Dice coefficient is $1.0000$ under common initialization versus $0.0105$ under independent initialization. Notably, $0.0105$ coincides with the random baseline expectation of $8/768 \approx 0.0104$ for independent rank-8 subspaces in 768 dimensions.

\begin{figure}[htbp!]
    \centering
    \includegraphics[width=\linewidth]{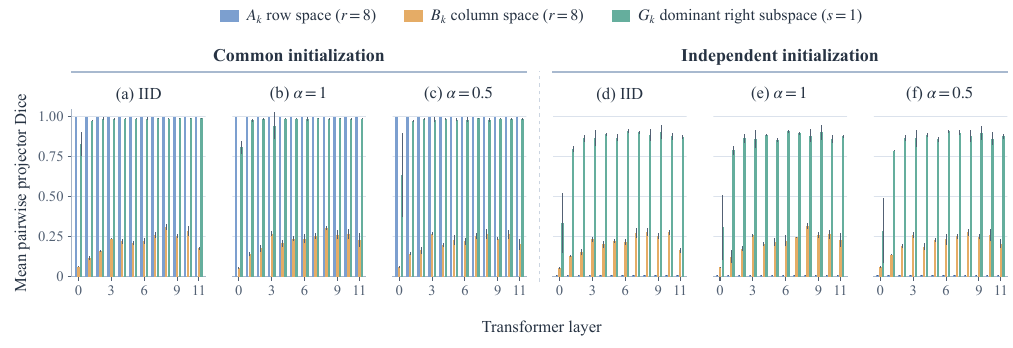}
    \caption{Query-projection subspace agreement in RoBERTa-base/RTE. For each initialization group, label heterogeneity increases from left to right, with $\alpha\in\{0.5,1.0\}$ controlling Dirichlet label skew.}
    \label{fig:app-query-subspaces}
\end{figure}
Figure~\ref{fig:app-query-subspaces} provides the query-projection
counterpart to Figure~\ref{fig:shared-object}. Both include all 12 layers,
with visibly lower independent-initialization $\bG$ agreement at layer 0.
At each layer, the three unordered client pairs are averaged within seed;
the plotted mean and standard deviation are computed across three seeds.
Pairs sharing a client are not treated as independent replications.

\begin{table}[htbp!]
\centering
\caption{Rank-$4$ $\bA$-rowspace Dice for  GPT-2, averaged over layers and query/value projections. $S_{11}$: same domain and task, $S_{10}$: same domain and different tasks, $S_{01}$: different domains and same task, $S_{00}$: different domains and tasks.}
\label{tab:gpt-init}
\small
\begin{tabular}{lcccc}
\toprule
Initialization & $S_{11}$ & $S_{10}$ & $S_{01}$ & $S_{00}$ \\
\midrule
Common & $0.9412$ & $0.8288$ & $0.9082$ & $0.8221$ \\
Independent & $0.0806$ & $0.0383$ & $0.0687$ & $0.0355$ \\
\bottomrule
\end{tabular}
\end{table}

The same initialization dependence is corroborated by the GPT-2 study (Table~\ref{tab:gpt-init}). Here, $S_{dt}$ identifies pairs with same-domain indicator $d$ and same-task indicator $t$. While independent initialization sharply reduces the complete $\bA$-rowspace overlap across all pairs, same-task pairs ($S_{11}, S_{01}$) retain slightly more overlap than different-task pairs. Thus, the loss of high absolute agreement in weight space should not be equated with the complete absence of task-specific geometry; rather, it demonstrates that raw parameter similarity is too fragile to serve as a robust medium for federated sharing.

\FloatBarrier
\subsection{Dominant and complete action-subspace agreement}
\label{app:dominant-components}
\begin{figure}[htbp!]
    \centering
    \includegraphics[width=\linewidth]{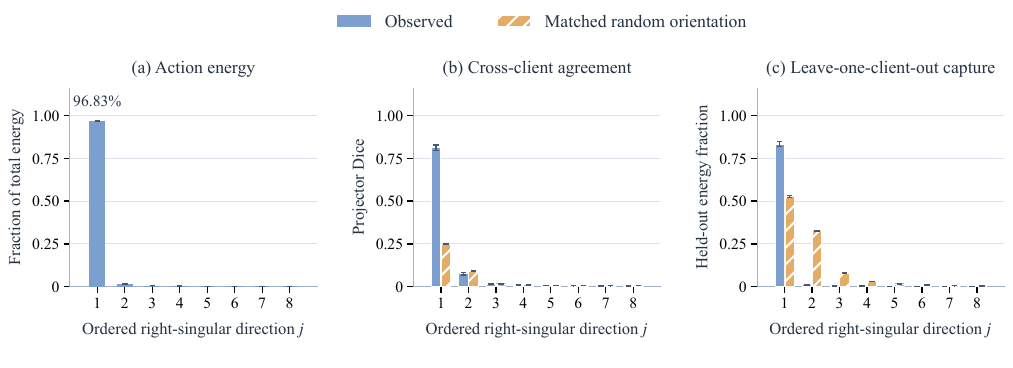}
    \caption{Ordered-direction diagnostics on RoBERTa-base/RTE with independent initialization and LoRA rank eight. Panels show (a) evaluation action energy fraction, (b) cross-client rank-one projector Dice, and (c) LOO action-energy capture. Metrics are averaged across partitions, layers, projections, and client pairs.}
    \label{fig:app-dominant-components}
\end{figure}

\paragraph{Agreement is concentrated in the leading direction.}
The leading direction captures $96.90\%$ and $96.83\%$ of the evaluation action energy under common and independent initialization, respectively, with a calibration/evaluation Dice coefficient of $0.9813$ in both cases. Under independent initialization, cross-client Dice agreement drops sharply as rank increases: $0.8113$ for the leading direction, $0.4536$ for the top-two subspace, and $0.1593$ for the full numerical-rank space (note that these rank-normalized overlaps do not measure captured energy). To examine individual directions, Figure~\ref{fig:app-dominant-components} evaluates all eight components separately. For each direction, we construct a leave-one-client-out (LOO) projector using the leading eigenvector of the mean calibration projector from the other two clients; LOO capture then measures the fraction of held-out evaluation action energy retained by this projector. As a control baseline, we replace the right singular basis of $(\bB_k^\top\bB_k)^{1/2}\bA_k$ with a uniformly sampled orthonormal basis while preserving its singular spectrum and local moments (averaged over 200 draws per seed, partition, layer, and projection). The leading direction substantially outperforms this control in both Dice agreement ($0.811$ vs.\ $0.248$) and LOO capture ($0.832$ vs.\ $0.528$), showing consistent positive gains across all seed means. In contrast, directions 2 through 8 perform below control on average across both metrics. These diagnostic findings confirm that cross-client agreement is strongly concentrated in the leading direction. We note that direction-wise LOO captures are non-additive and serve as diagnostic insights rather than guarantees of downstream aggregation performance.

\FloatBarrier
\subsection{Task, domain, and layer dependence in GPT-2}
\label{app:task-layer-diagnostics}

For Figure~\ref{fig:gpt2-task-structure}, $S_{dt}$ denotes the mean
fixed-top-$1$ action-subspace Dice in each of the four pair classes.
There are 4 pairs in $S_{11}$ and 8 each in $S_{10}$, $S_{01}$, and
$S_{00}$. We first average within each class, then compute the balanced
factorial contrasts
\begin{equation}
    C_{\rm task}=\tfrac12(S_{11}+S_{01}-S_{10}-S_{00}),\qquad
    C_{\rm domain}=\tfrac12(S_{11}+S_{10}-S_{01}-S_{00}),
\end{equation}
which give each pair class equal weight. A positive $C_{\rm task}$
indicates greater agreement between same-task clients after averaging
equally over same-domain and different-domain pairs. The comparison
$S_{01}-S_{10}=C_{\rm task}-C_{\rm domain}$ asks the stronger question
of whether task-related differences exceed domain-related differences.
Its sign describes their relative strength; a negative value can coexist
with a positive task contrast.

\begin{table}[!ht]
\centering
\caption{Task and domain organization of the fixed rank-$1$ dominant right direction of $\bG_k$. Mean values and $95\%$ $t$ intervals (in brackets) are averaged over query and value projections.}
\label{tab:gpt-task-structure}
\small
\begin{tabular}{llccc}
\toprule
Initialization & Tokens & $C_{\rm task}$ & $C_{\rm domain}$ & $S_{01}-S_{10}$ \\
\midrule
Common & Instruction & $.224 [ .207, .240]$ & $.008 [ .001, .016]$ & $.215 [ .195, .235]$ \\
Common & Content     & $.212 [ .202, .222]$ & $.047 [ .042, .052]$ & $.165 [ .153, .176]$ \\
Common & Decision    & $.242 [ .221, .263]$ & $.023 [ .017, .030]$ & $.218 [ .196, .241]$ \\
Independent & Instruction & $.201 [ .190, .213]$ & $.011 [ .003, .018]$ & $.191 [ .178, .203]$ \\
Independent & Content     & $.201 [ .192, .211]$ & $.039 [ .033, .045]$ & $.162 [ .153, .171]$ \\
Independent & Decision    & $.237 [ .222, .252]$ & $.023 [ .017, .029]$ & $.214 [ .198, .230]$ \\
\bottomrule
\end{tabular}
\end{table}

At $s=1$, task contrasts are positive and exceed domain contrasts under
both initialization conditions in all three token regions
(Table~\ref{tab:gpt-task-structure}).
Different-task pairs nevertheless retain mean Dice of approximately
$0.65$--$0.67$. Task dependence therefore appears as a difference within
partly shared geometry, not as orthogonality between tasks. In the
layerwise curves, this difference is most pronounced in later content
and decision layers. Early decision layers remain highly aligned even
across tasks. These depth-dependent patterns motivate a layerwise sharing
criterion; the effects of actually disabling guidance are tested separately
in Appendix~\ref{app:layer-selection}.

\begin{table}[htbp!]
\centering
\caption{Task-minus-domain contrasts in GPT-2 across projections and retained ranks, averaged over replicates and 12 layers. Entries report $S_{01}-S_{10}=C_{\rm task}-C_{\rm domain}$, with negative values indicating a larger domain contrast. $Q$ and $V$ denote query and value.}
\label{tab:app-gpt-rank-projection}
\small
\begin{tabular}{lllrrrr}
\toprule
Initialization & Tokens & Projection & $s=1$ & $s=2$ & $s=3$ & $s=4$ \\
\midrule
Common & Instruction & Q & $.211$ & $.246$ & $.288$ & $.318$ \\
       &             & V & $.220$ & $.287$ & $.328$ & $.356$ \\
       & Content     & Q & $.126$ & $.097$ & $.016$ & $-.061$ \\
       &             & V & $.204$ & $.150$ & $.016$ & $-.067$ \\
       & Decision    & Q & $.210$ & $.188$ & $.129$ & $.083$ \\
       &             & V & $.227$ & $.207$ & $.144$ & $.082$ \\
\midrule
Independent & Instruction & Q & $.191$ & $.138$ & $.132$ & $.129$ \\
            &             & V & $.190$ & $.137$ & $.123$ & $.117$ \\
            & Content     & Q & $.126$ & $.079$ & $.047$ & $.028$ \\
            &             & V & $.199$ & $.123$ & $.082$ & $.052$ \\
            & Decision    & Q & $.210$ & $.156$ & $.121$ & $.092$ \\
            &             & V & $.219$ & $.203$ & $.148$ & $.107$ \\
\bottomrule
\end{tabular}
\end{table}
\paragraph{Projection and retained-rank checks.}
Table~\ref{tab:app-gpt-rank-projection} separates query and value and
varies the retained analysis dimension $s$ on the same rank-$4$ adapters.
The balanced task contrast $C_{\rm task}$, averaged over query and value,
remains positive in every token region under both initialization schemes
for all $s\in\{1,2,3,4\}$. Task-conditioned differences therefore remain
detectable in larger retained subspaces in this experiment.
Under independent initialization, $S_{01}-S_{10}$ remains positive for
both projections at every tested rank, with all seed-level intervals
excluding zero; $s=1$ gives the largest contrast in each token region.
Under common initialization, the full content-token subspace ($s=4$)
has a larger domain contrast than task contrast, yielding negative
$S_{01}-S_{10}$ for both projections. Its projection-averaged task
contrast nevertheless remains positive: $C_{\rm task}=0.190$ with a
$95\%$ seed-level $t$ interval of $[0.183,0.197]$. Thus, the reversal
concerns the relative strength of task and domain structure, while
task dependence remains present. The relative contributions vary with
initialization and retained rank.

\FloatBarrier

\section{Personalized GLUE: Training and Evaluation Protocol}
\label{app:glue-protocol}

\subsection{Tasks, client partitions, and common settings}

We evaluate RTE, SST-2, QNLI, MNLI, and QQP, with a separate three-client
federation for each task. MNLI-m and MNLI-mm are reported separately,
giving six evaluation splits. Training examples are partitioned by a
label-skew Dirichlet distribution with
$\alpha\in\{0.5,0.75,1.0\}$; smaller $\alpha$ produces stronger label
heterogeneity. Each client's validation partition follows the empirical
label mixture of its training partition. For each task, $\alpha$, and seed,
all methods use the same client training and validation indices.Table~\ref{tab:glue-training-protocol} gives the common settings for the
main comparison and the GLUE ablation and rank studies. The latter vary
only the settings identified in Appendices~\ref{app:core-ablation}--\ref{app:rank-sensitivity}.

\begin{table}[!htbp]
    \centering
    \caption{Common training settings for personalized GLUE. FedSAIL-specific
    settings also apply to its message and rank ablations unless stated otherwise.}
    \label{tab:glue-training-protocol}
    \small
    \setlength{\tabcolsep}{5pt}
    \renewcommand{\arraystretch}{1.08}
    \begin{tabular}{@{}p{0.27\linewidth}p{\dimexpr0.73\linewidth-2\tabcolsep\relax}@{}}
        \toprule
        Setting & Configuration \\
        \midrule
        Backbone and adapters & RoBERTa-large; query and value projections in
        all $24$ layers; default $r_{\rm L}=8$, $\alpha_{\rm LoRA}=16$,
        LoRA dropout $0.05$. \\
        Initialization & Independent client $\bA_k$, zero $\bB_k$;
        identical initial classification heads, trained locally. \\
        Data and batching & Full task training/validation sets;
        maximum length $128$; batch and micro-batch size $128$. \\
        Optimization & SGD with CUDA AMP; $500$ rounds;
        $10$ local steps per client and phase per round. \\
        Main-comparison seeds & $123$, $124$, and $125$. \\
        Evaluation & Personalized validation every $25$ rounds;
        no global-model evaluation. \\
        \midrule
        FedSAIL guidance & Shared rank $s=1$, logged penalty $\beta=1$;
        $50$ local-only warmup rounds, then $450$ guided rounds;
        layer selection disabled. \\
        Input moments & Estimated from the first $128$ examples of each
        client's training partition before round $1$;
        shrinkage $\gamma=0.05$; frozen throughout training. \\
        \bottomrule
    \end{tabular}
\end{table}

\subsection{Method-specific settings and scale convention}

In Table~\ref{tab:glue-method-protocol} we distinguish the communication and
optimization rules. All classification
heads remain client-local. Baselines aggregate uniformly from the first
round. Our FedSAIL approach starts sharing after warmup and weights its trace-normalized
spectral messages by the number of local tokens used to estimate the moments.

\begin{table}[!htbp]
    \centering
    \caption{Method-specific GLUE settings. Steps are local steps per client
    and round; FedDPA-LoRA uses separate global and local phases.}
    \label{tab:glue-method-protocol}
    \small
    \setlength{\tabcolsep}{4pt}
    \renewcommand{\arraystretch}{1.08}
    \begin{tabular*}{\linewidth}{@{\extracolsep{\fill}}llcc@{}}
        \toprule
        Method & Federated exchange & Learning rate & Steps \\
        \midrule
        LoRA & Local $\bA_k$ and $\bB_k$ & $0.005$ & $10$ \\
        FFA-LoRA & Average $\bB_k$; freeze local $\bA_k$ & $0.005$ & $10$ \\
        FedSA-LoRA & Average $\bA_k$; keep $\bB_k$ local & $0.005$ & $10$ \\
        FedDPA-LoRA & Average global factors; retain local branch & $0.005$ & $10+10$ \\
        FedSAIL & Share action spectra and consensus bases & $0.005$ & $10$ \\
        \bottomrule
    \end{tabular*}
\end{table}

We shall note that FedDPA-LoRA has separate rank-$8$ global and local adapters and uses a fixed
global/local prediction mixture of $0.5/0.5$. Its adapter capacity and
scheduled local-step budget are twice those of the single-branch methods.
Another thing that worth attention is that for FFA-LoRA, freezing independently initialized $\bA_k$ differs from its
canonical shared-$\bA$ setting, so that the reported performance may not achieve its best possible performance.

\FloatBarrier

\section{Shared-Message Ablation on Personalized GLUE}
\label{app:core-ablation}

\paragraph{Controlled comparison.}
To understand exactly what makes the shared consensus effective, we isolate the components of our transmitted message. Specifically, we ask whether the performance gains stem from incorporating the local input moments ($\widehat\bSigma_k^{1/2}$), the up-projection ($\bB_k$), or their combination. In our experiments, we compare the rank-$1$ right spectral messages extracted from four variants: $\bA_k$, the full raw update $\bDelta_k$, $\bA_k\widehat\bSigma_k^{1/2}$, and our full input-aware action matrix $\widehat\bG_k=\bDelta_k\widehat\bSigma_k^{1/2}$. Besides, to ensure a strict comparison of the messages themselves, we hold the local optimization objective constant across all variants. We evaluate these variants using seed $123$ across all three Dirichlet heterogeneity settings. To rule out confounding initialization effects, all configurations are strictly matched from step zero, sharing identical client partitions, data-loader states, and warmup trajectories up to round $50$.

\paragraph{Results and interpretation.}
As shown in Table~\ref{tab:glue-message-ablation-full},  incorporating the input-moment weighting improves performance across all $18$ task-by-heterogeneity configurations. Without $\widehat\bSigma_k^{1/2}$, both unweighted messages ($\bA_k$ and $\bDelta_k$) suffer severe performance degradation, averaging only $66.68\%$. Notably, even though sharing the full raw update $\bDelta_k$ resolves the non-uniqueness (reparameterization ambiguity) of individual LoRA factors, it still performs poorly under the fixed action-based regularizer. This demonstrates a key insight: merely fixing factorization ambiguity is insufficient. To provide meaningful cross-client guidance, the shared geometric consensus must be explicitly anchored in the local input distribution. We also observe that extracting the shared subspace from $\bA_k\widehat\bSigma_k^{1/2}$ yields an overall mean ($88.92\%$) that is only marginally lower than using the full action matrix $\widehat\bG_k$ ($89.02\%$). Consistent with our discussion in the main text, this minor empirical gap suggests that including the output factor $\bB_k$ in the shared message is not strictly necessary for achieving high predictive performance on these tasks. However, our preference for the full action matrix $\widehat\bG_k$ is mathematically motivated. Unlike the right subspace of $\bA_k\widehat\bSigma_k^{1/2}$, $\widehat\bG_k$ is inherently invariant to invertible LoRA reparameterizations. This coordinate-free property makes it a fundamentally more robust and principled geometric object to share across heterogeneous clients. Finally, we note that these single-seed results specifically assess the composition of the shared message; to maintain a strictly controlled comparison, all evaluated variants still retain both $\bB_k$ and the input moment in their local optimization objective.

\begin{table}[!htbp]
    \centering
    \caption{Ablation of shared-message on personalized GLUE. We report validation accuracy (\%) over $500$ rounds.}
    \label{tab:glue-message-ablation-full}
    \scriptsize
    \setlength{\tabcolsep}{2pt}
    \begin{tabular*}{\linewidth}{@{\extracolsep{\fill}}clccccccc@{}}
        \toprule
        $\alpha$ & Shared message & RTE & SST-2 & QNLI &
        MNLI-m & MNLI-mm & QQP & Avg. \\
        \midrule
        $0.5$ & $\bA_k$ & $74.73$ & $81.42$ & $70.49$ & $48.04$ & $49.14$ & $69.11$ & $65.49$ \\
        & $\bDelta_k$ & $74.37$ & $81.65$ & $70.55$ & $48.13$ & $49.15$ & $68.56$ & $65.40$ \\
        & $\bA_k\widehat\bSigma_k^{1/2}$ & $\mathbf{85.20}$ & $95.30$ & $93.25$ & $87.45$ & $87.39$ & $\mathbf{86.01}$ & $89.10$ \\
        & $\widehat\bG_k$ & $\mathbf{85.20}$ & $\mathbf{95.53}$ & $\mathbf{93.37}$ & $\mathbf{87.56}$ & $\mathbf{87.55}$ & $85.92$ & $\mathbf{89.19}$ \\
        \addlinespace
        $0.75$ & $\bA_k$ & $63.54$ & $76.72$ & $69.25$ & $62.65$ & $63.10$ & $68.35$ & $67.27$ \\
        & $\bDelta_k$ & $61.73$ & $78.67$ & $69.34$ & $62.69$ & $63.13$ & $68.87$ & $67.40$ \\
        & $\bA_k\widehat\bSigma_k^{1/2}$ & $\mathbf{83.39}$ & $95.41$ & $93.04$ & $89.04$ & $89.25$ & $85.35$ & $89.25$ \\
        & $\widehat\bG_k$ & $83.03$ & $\mathbf{95.53}$ & $\mathbf{93.26}$ & $\mathbf{89.06}$ & $\mathbf{89.37}$ & $\mathbf{85.67}$ & $\mathbf{89.32}$ \\
        \addlinespace
        $1.0$ & $\bA_k$ & $70.40$ & $72.25$ & $70.25$ & $59.36$ & $59.43$ & $71.95$ & $67.27$ \\
        & $\bDelta_k$ & $70.04$ & $72.48$ & $70.14$ & $59.36$ & $59.43$ & $71.99$ & $67.24$ \\
        & $\bA_k\widehat\bSigma_k^{1/2}$ & $77.98$ & $95.99$ & $\mathbf{93.06}$ & $88.51$ & $\mathbf{88.52}$ & $86.40$ & $88.41$ \\
        & $\widehat\bG_k$ & $\mathbf{78.34}$ & $\mathbf{96.44}$ & $92.84$ & $\mathbf{88.65}$ & $88.44$ & $\mathbf{86.63}$ & $\mathbf{88.56}$ \\
        \bottomrule
    \end{tabular*}
\end{table}

\FloatBarrier

\section{Shared and Local Rank Sensitivity on Personalized GLUE}
\label{app:rank-sensitivity}

A defining architectural advantage of FedSAIL over traditional parameter-averaging methods is its ability to decouple the communication payload from the client's local adapter capacity. In this section, we provide the detailed communication accounting and empirical sensitivity analysis for the shared rank ($s$) and local rank ($r_{\rm L}$).

\paragraph{Communication accounting.}
At each adapted module, a client uploads an orthonormal basis of size $1024 \times s$, $s$ eigenvalues, and a single integer token count, and subsequently receives a $1024 \times s$ global consensus basis. Assuming float32 precision for bases and eigenvalues and int32 for the count, the total bidirectional payload per run is:
\begin{equation}
    C_{\rm total}(s)
    = 3 \text{ (clients)} \times 48 \text{ (modules)} \times 450 \text{ (rounds)} \times 4\,(2049s+1)\quad\text{bytes}.
    \label{eq:rank-communication}
\end{equation}
Each client has $98{,}304r_{\rm L}$ trainable LoRA parameters across all $24$ layers. Under FedSAIL, the communication payload (Equation~\ref{eq:rank-communication}) scales exclusively with $s$ and remains completely independent of $r_{\rm L}$. For a fixed $s=1$, the total communication is $0.4949$ GiB, strictly governing the network overhead regardless of how large the local adapters grow.

\paragraph{Experimental setup.}
To empirically evaluate this decoupling, we conduct two complementary sweeps around the default configuration of $(r_{\rm L},s)=(8,1)$ following the protocol before. 
\begin{enumerate}
    \item \textbf{Shared-rank sweep}: We vary $s \in \{1,2,4,8\}$ while fixing $r_{\rm L}=8$ to examine the communication-accuracy trade-off.
    \item \textbf{Local-rank sweep}: We vary $r_{\rm L} \in \{2,4,8,16\}$ while fixing $s=1$ to examine local capacity scaling under a strict communication budget.
\end{enumerate}
To maintain isolation between variables, we fix the effective unscaled-factor penalty coefficient at $\beta_{\rm eff}=4$ by setting the LoRA scaling factor $\alpha_{\rm LoRA}=2r_{\rm L}$. Neither the learning rate nor the penalty weight is retuned across either sweep. Tables~\ref{tab:glue-shared-rank} and \ref{tab:glue-adapter-rank} report the personalized validation accuracy over $500$ rounds. Replicated entries use seeds $123$, $124$, and $125$; we compute the six-split mean within each seed before calculating its sample standard deviation across seeds.

\begin{table}[htbp!]
    \centering
    \caption{Shared-rank sensitivity with fixed local rank $r_{\rm L}=8$. 
    We report personalized validation accuracy and total communication cost, with accuracy averaged across the six splits.}
    \label{tab:glue-shared-rank}
    \footnotesize
    \setlength{\tabcolsep}{2pt}
    \renewcommand{\arraystretch}{1.10}
    \begin{tabular*}{\linewidth}{@{\extracolsep{\fill}}cccccccccc@{}}
        \toprule
        $\alpha$ & $s$ & RTE & SST-2 & QNLI & MNLI-m & MNLI-mm & QQP
        & Avg. & \shortstack{Comm. (GiB)} \\
        \midrule
        $0.5$ & $1$ & $80.87_{\pm7.50}$ & $96.37_{\pm1.17}$ & $93.06_{\pm0.57}$ & $88.98_{\pm1.54}$ & $89.07_{\pm1.75}$ & $89.80_{\pm5.39}$ & $89.69_{\pm1.28}$ & $0.4949$ \\
         & $2$ & $80.39_{\pm5.53}$ & $96.56_{\pm1.30}$ & $93.56_{\pm0.57}$ & $89.12_{\pm1.60}$ & $89.23_{\pm1.80}$ & $90.11_{\pm5.34}$ & $89.83_{\pm1.08}$ & $0.9895$ \\
         & $4$ & $80.51_{\pm7.82}$ & $96.64_{\pm1.32}$ & $93.40_{\pm0.31}$ & $89.37_{\pm1.50}$ & $89.45_{\pm1.81}$ & $90.07_{\pm5.39}$ & $89.91_{\pm1.48}$ & $1.9787$ \\
         & $8$ & $80.75_{\pm8.08}$ & $96.60_{\pm1.30}$ & $93.37_{\pm0.59}$ & $89.34_{\pm1.53}$ & $89.41_{\pm1.80}$ & $90.19_{\pm5.44}$ & $89.94_{\pm1.50}$ & $3.9573$ \\
        \midrule
        $0.75$ & $1$ & $77.26_{\pm5.05}$ & $95.64_{\pm0.20}$ & $93.23_{\pm0.61}$ & $88.61_{\pm1.53}$ & $88.84_{\pm1.60}$ & $89.05_{\pm4.69}$ & $88.77_{\pm0.56}$ & $0.4949$ \\
         & $2$ & $76.90_{\pm5.32}$ & $95.80_{\pm0.52}$ & $93.84_{\pm1.04}$ & $88.72_{\pm1.49}$ & $88.85_{\pm1.60}$ & $89.10_{\pm4.79}$ & $88.87_{\pm0.49}$ & $0.9895$ \\
         & $4$ & $77.86_{\pm3.62}$ & $95.68_{\pm0.46}$ & $93.78_{\pm1.06}$ & $89.04_{\pm1.44}$ & $89.12_{\pm1.55}$ & $89.43_{\pm4.56}$ & $89.15_{\pm0.18}$ & $1.9787$ \\
         & $8$ & $77.86_{\pm3.71}$ & $95.68_{\pm0.18}$ & $93.65_{\pm1.32}$ & $89.12_{\pm1.54}$ & $89.10_{\pm1.66}$ & $89.40_{\pm4.60}$ & $89.13_{\pm0.18}$ & $3.9573$ \\
        \midrule
        $1.0$ & $1$ & $81.95_{\pm3.61}$ & $96.25_{\pm1.10}$ & $93.28_{\pm0.41}$ & $88.40_{\pm1.40}$ & $88.52_{\pm1.35}$ & $88.72_{\pm2.11}$ & $89.52_{\pm1.14}$ & $0.4949$ \\
         & $2$ & $81.71_{\pm2.71}$ & $96.29_{\pm1.09}$ & $93.87_{\pm0.60}$ & $88.47_{\pm1.29}$ & $88.70_{\pm1.49}$ & $88.87_{\pm2.13}$ & $89.65_{\pm0.98}$ & $0.9895$ \\
         & $4$ & $80.63_{\pm2.71}$ & $96.02_{\pm0.81}$ & $93.79_{\pm0.61}$ & $88.67_{\pm1.31}$ & $88.75_{\pm1.50}$ & $88.96_{\pm2.11}$ & $89.47_{\pm1.03}$ & $1.9787$ \\
         & $8$ & $80.75_{\pm2.94}$ & $96.06_{\pm1.10}$ & $93.81_{\pm0.60}$ & $88.64_{\pm1.44}$ & $88.78_{\pm1.47}$ & $88.84_{\pm1.97}$ & $89.48_{\pm1.04}$ & $3.9573$ \\

        \bottomrule
    \end{tabular*}
\end{table}

\paragraph{Decoupling geometric alignment from local personalization.}
The results from the two sweeps illustrate the core mechanism of FedSAIL: structural alignment and task fitting utilize different forms of capacity. As shown in Table~\ref{tab:glue-shared-rank}, increasing the shared rank $s$ yields only small, non-monotonic changes in accuracy. Expanding $s$ from $1$ to $4$ quadruples the communication payload (from $0.49$ GiB to $1.98$ GiB) but results in heavily diminishing returns in predictive performance. This pattern is qualitatively consistent with our diagnostic findings in Section~\ref{sec:action-matrix}, where cross-client agreement is concentrated in the leading direction. These results support $s=1$ as a communication-efficient default, while larger shared ranks offer modest, task-dependent gains at increased communication cost.

\begin{table}[htbp!]
    \centering
    \caption{Local-rank sensitivity with fixed shared rank $s=1$. We report six-split validation accuracy (\%, mean $\pm$ sample std.) and per-client LoRA parameter counts. Total communication remains fixed at $0.4949$ GiB per run.}
    \label{tab:glue-adapter-rank}
    \small
    \setlength{\tabcolsep}{4pt}
    \renewcommand{\arraystretch}{1.10}
    \begin{tabular*}{\linewidth}{@{\extracolsep{\fill}}ccccc@{}}
        \toprule
        $r_{\rm L}$ & Params. (M) & $\alpha=0.5$ & $\alpha=0.75$
        & $\alpha=1.0$ \\
        \midrule
        $2$ & $0.197$ & $88.74_{\pm 1.43}$ & $87.92_{\pm 0.55}$ & $88.65_{\pm 1.41}$ \\
        $4$ & $0.393$ & $89.21_{\pm 1.51}$ & $88.24_{\pm 0.81}$ & $89.10_{\pm 1.09}$ \\
        $8$ & $0.786$ & $89.69_{\pm 1.28}$ & $88.77_{\pm 0.56}$ & $89.52_{\pm 1.14}$ \\
        $16$ & $1.573$ & $89.84_{\pm 1.43}$ & $89.22_{\pm 0.29}$ & $89.61_{\pm 0.95}$ \\
        \bottomrule
    \end{tabular*}
\end{table}
Conversely, Table~\ref{tab:glue-adapter-rank} shows that additional local capacity can improve accuracy without increasing the message size, although the gains depend on the heterogeneity setting. Clients retain the necessary degrees of freedom ($r_{\rm L} > s$) to model task-specific residuals, while collaborating strictly within a low-dimensional ($s=1$) consensus geometry.

\FloatBarrier

\section{Real-DDDT Layer Selection and Communication}
\label{app:layer-selection}

\subsection{Benchmark and Training Protocol}
To evaluate layer selection, we construct two disjoint client replicas for each of the following datasets: IMDb (sentiment), Amazon (polarity), QQP (paraphrase), and QNLI (entailment). Within each replica, IMDb is paired with QQP, and Amazon with QNLI, yielding four disjoint different-domain/different-task dyads. Guidance-basis aggregation is strictly restricted to each dyad, while both LoRA factors remain entirely local. The geometric selector pools their subspace alignments to compute one common layer-sharing mask per training seed. Table~\ref{tab:dddt-protocol} details the dataset allocation and hyperparameter configurations.

\begin{table}[!htbp]
    \centering
    \caption{DDDT data, training, and evaluation protocol.}
    \label{tab:dddt-protocol}
    \small
    \setlength{\tabcolsep}{4pt}
    \renewcommand{\arraystretch}{1.08}
    \begin{tabular}{@{}p{0.24\linewidth}p{\dimexpr0.76\linewidth-2\tabcolsep\relax}@{}}
        \toprule
        Component & Setting \\
        \midrule
        Clients and splits & Eight clients; per client: $1000$ training,
        $256$ calibration, $244$ validation, and $500$ final-test examples.
        All splits are label-balanced and disjoint across splits and replicas. \\
        Data sources & Training/calibration: official training pools.
        Validation/final test: IMDb/Amazon test pools and labeled QQP/QNLI
        validation pools; no GLUE test-server evaluation. \\
        Seeds & Data partition: $2027$.
        Training seeds: $83,97,109$. \\
        Backbone and adapters & Frozen RoBERTa-large and MLM head;
        query/value LoRA in all $24$ layers, $r_{\rm L}=4$,
        $\alpha_{\rm LoRA}=16$, dropout $0.05$.
        Independent $\bA_k$ initialization and $\bB_k=0$. \\
        Optimization & AdamW, learning rate $2\times10^{-4}$, weight decay
        $0$, gradient clipping $1$, batch size $16$, CUDA AMP;
        $10$ local iterations per client per round.
        Optimizer/scaler states remain local across rounds. \\
        Training schedule & $64$ rounds: $32$ local warmup rounds and $32$ guided rounds,
        with shared rank $s=1$ and guidance coefficient $\beta=1$. \\
        Inputs and prediction & Maximum $128$ tokens, preserving instructions,
        field markers, decision mask, and content heads/tails.
        Mask-token verbalizers: negative/positive (sentiment),
        different/same (paraphrase), and yes/no (entailment). \\
        Evaluation & Terminal round-$64$ checkpoint. Average accuracy over
        eight clients within each seed, then report mean and sample SD over
        three seeds. Task columns average two replicas and three seeds. \\
        \bottomrule
    \end{tabular}
\end{table}
To prevent data leakage and ensure rigorous evaluation, review-body and question-identity deduplication are applied to control direct overlap. Equal final-test sizes are maintained to strictly align client-macro and sample-micro accuracy metrics; sentiment tasks contribute half the weight, while paraphrase and entailment each contribute a quarter.

\subsection{Warmup-Based Layer Selection}
During local warmup ($\beta=0$), we collect snapshots at rounds
$t\in\{24,28,32\}$. For client $k$, layer $\ell\in\{0,\ldots,23\}$,
query/value projection $a\in\{Q,V\}$, and token regime
$z\in\mathcal Z=\{\text{instruction},\text{content},\text{decision}\}$,
let $\mathbf q_{k\ell azt}$ be a unit leading right singular vector of
$\bG_{k\ell azt}=\bB_{k\ell a}^{t}\bA_{k\ell a}^{t}
\bSigma_{k\ell zt}^{1/2}$.
Here $\bSigma_{k\ell zt}$ is the uncentered second moment of local
calibration inputs in regime $z$ at layer $\ell$ and round $t$, computed
without shrinkage and with negative eigenvalues clipped to zero.
For each of the four dyads $d\in\{1,2,3,4\}$ described above,
$i_d$ and $j_d$ denote its two client indices.
To ensure stability against stochastic training noise, we define the layer agreement score $s_\ell$ using a temporal median over the collected snapshots:
\begin{equation}
    s_\ell=\operatorname{median}_{t\in\{24,28,32\}}
    \left[\frac{1}{24}\sum_{d=1}^{4}\sum_{a\in\{Q,V\}}\sum_{z\in\mathcal Z}
    \left|\mathbf{q}_{i_d\ell azt}^{\top}\mathbf{q}_{j_d\ell azt}\right|^2\right],
    \qquad
    \mathcal S_\tau=\{\ell:s_\ell\geq\tau\}.
    \label{eq:dddt-selector}
\end{equation}
Each snapshot averages $4\times2\times3=24$ rank-one Projector Dice values.
This threshold $\tau$ applies to the pooled score rather than individual dyads. Once the mask $\mathcal S_\tau$ is computed at round $32$, it remains fixed. Notably, we impose no contiguity constraints or minimum layer requirements. For instance, at $\tau=0.6$, the selector naturally excludes the final six layers in every random seed, consistently identifying that deeper layers encode representations too task-specific to benefit from federated sharing.

Following selection, all arms branch from identical round-$32$ optimizer and scaler states. Selected layers use the subspace penalty defined in \eqref{eq:local-objective} with $\beta=1$, while unselected layers continue purely local adaptation using only the task loss.

\subsection{Threshold Sensitivity}

Table~\ref{tab:dddt-layer-selection} compares full-layer guidance against four different thresholds ($\tau$) applied to the exact same warmup geometry. Increasing $\tau$ imposes a stricter geometric consensus requirement, restricting sharing to layers with robust cross-client agreement.
\begin{table}[!htbp]
    \centering
    \caption{FedSAIL threshold sensitivity on DDDT. We report round-$64$ accuracy (\%, mean $\pm$ sample std.) and total communication cost, including selector snapshots and mask delivery.}
    \label{tab:dddt-layer-selection}
    \small
    \setlength{\tabcolsep}{3pt}
    \begin{tabular*}{\linewidth}{@{\extracolsep{\fill}}lccccccc@{}}
        \toprule
        FedSAIL variant & $\tau$ & IMDb & Amazon & QQP & QNLI & Avg. & Comm. \\
        \midrule
        Full & -- & $93.33$ & $95.93$ & $\mathbf{83.27}$ & $87.60$ & $90.033_{\pm0.225}$ & $96.094$ \\
        \midrule
        Selection & $0.4$ & $93.20$ & $95.70$ & $83.07$ & $88.07$ & $90.008_{\pm0.644}$ & $104.282$ \\
        Selection & $0.5$ & $93.23$ & $96.17$ & $82.37$ & $\mathbf{88.37}$ & $90.033_{\pm0.284}$ & $89.601$ \\
        Selection & $0.6$ & $93.57$ & $96.37$ & $83.13$ & $87.47$ & $\mathbf{90.133}_{\pm0.038}$ & $77.589$ \\
        Selection & $0.7$ & $\mathbf{93.77}$ & $\mathbf{96.40}$ & $81.97$ & $87.50$ & $89.908_{\pm0.800}$ & $\mathbf{58.904}$ \\
        \bottomrule
    \end{tabular*}
\end{table}

The sensitivity sweep clearly demonstrates the efficacy of geometry-guided selection. At $\tau=0.6$, restricting guidance exclusively to the selected layers reduces the total communication payload by $19.26\%$ while retaining predictive performance entirely comparable to full-layer guidance (a minimal difference of $+0.100$ percentage points, with a $95\%$ $t$ interval of $[-0.530,0.730]$). This validates our core hypothesis: layers with divergent action subspaces should be left entirely local, freeing them from negative transfer (unhelpful cross-client penalties) while saving network bandwidth.

Furthermore, this sweep serves as a comprehensive sensitivity analysis for our system. We observe a clear trade-off surface: while $\tau=0.7$ minimizes communication maximally, it begins to degrade mean accuracy (particularly on QQP) due to overly aggressive isolation. Conversely, a highly permissive threshold like $\tau=0.4$ retains so many layers that the overhead of the selector mechanism outpaces the communication saved. 

\subsection{Communication Accounting}

Using the per-module payload convention in
Appendix~\ref{app:rank-sensitivity}, each active query/value module at
$s=1$ communicates $4(2049+1)$ bytes per client per guided round.
For this DDDT protocol, eight clients, two modules per layer, and $32$
guided rounds give $4{,}198{,}400L$ bytes for $L$ active layers.
The additional selector cost consists of three snapshots over all
$24$ layers and three token regimes,
$3\times8\times24\times2\times3\times1026\times4=14{,}183{,}424$ bytes,
plus $24$ bytes to deliver a $24$-bit mask to each of the eight clients.
Thus,
\begin{equation}
    C_{\rm full}=4{,}198{,}400\times24,\qquad
    C_{\rm selection}(L)=4{,}198{,}400L+14{,}183{,}448
    \quad\text{bytes}.
    \label{eq:dddt-communication}
\end{equation}
Full-layer FedSAIL requires no selector. At this training length,
selection must retain at most $20$ layers to offset its overhead.
Table~\ref{tab:dddt-layer-selection} reports the mean per-run cost over
three seeds in MiB ($2^{20}$ bytes), charging the full selector cost to
each threshold without amortization across the sweep.

These are logical payloads counted for every client delivery, excluding
initial model distribution, headers, serialization, checkpoints,
optimizer states, and evaluation traffic. The reported deployment cost
includes the three snapshots required by the selection rule. The research
pilot recorded five additional diagnostic snapshots; charging all eight
adds $22.544$ MiB per selection run. Under this research-logging allocation,
$\tau=0.6$ costs $100.133$ MiB rather than $77.589$ MiB and does not save
communication relative to full-layer FedSAIL. The savings reported in
the main text therefore refer to the three-snapshot selection protocol.

\FloatBarrier

\section{Code Availability and Computational Resources}
\label{app:code-resources}
\paragraph{Computational resources.}
Table~\ref{tab:compute-resources} summarizes the primary compute server and
benchmark training environment. GLUE and DDDT runs simulate
federated clients within single-GPU jobs; independent jobs are scheduled
across the available GPUs. Both protocols use CUDA automatic mixed
precision, with training settings given in
Appendices~\ref{app:glue-protocol} and~\ref{app:layer-selection}.

\begin{table}[!htbp]
    \centering
    \caption{Primary compute server and benchmark training environment.}
    \label{tab:compute-resources}
    \small
    \setlength{\tabcolsep}{5pt}
    \renewcommand{\arraystretch}{1.08}
    \begin{tabular}{@{}p{0.23\linewidth}p{\dimexpr0.77\linewidth-2\tabcolsep\relax}@{}}
        \toprule
        Component & Specification \\
        \midrule
        GPUs & $4\times$ NVIDIA GeForce RTX 3090, $24$ GiB memory each \\
        CPUs & $2\times$ Intel Xeon Gold 5218R, $40$ physical cores in total \\
        Host memory & Approximately $504$ GiB (OS-reported) \\
        Operating system & Ubuntu 22.04.5 LTS \\
        Training runtime & Python 3.9.21; PyTorch 2.5.1 with CUDA 12.4 \\
        Libraries & Transformers 4.42.3; Datasets 2.20.0 \\
        \bottomrule
    \end{tabular}
\end{table}

\paragraph{Anonymous code release.}
The core implementation of FedSAIL is available at
\begin{center}
    \url{https://anonymous.4open.science/r/FedSAIL-4A7C/}.
\end{center}
The repository implements input-moment estimation and shrinkage, compact
spectral messages, server-side subspace aggregation, the local action
residual penalty, layer selection, and communication accounting.
Synthetic CPU tests check these components against dense reference
computations, including reparameterization invariance and penalty gradients.
Experimental settings and evaluation conventions are documented
in the preceding appendices.

\paragraph{Installation and verification.}
The standalone package requires Python $3.10$ or later and PyTorch.
From the repository root, the following commands install the package
and run its numerical tests without downloading models or datasets:
\begin{verbatim}
python -m pip install -e ".[test]"
PYTHONDONTWRITEBYTECODE=1 python -m pytest -p no:cacheprovider
\end{verbatim}

\end{document}